\documentclass[10pt]{article}
\usepackage[utf8]{inputenc}
\usepackage[T1]{fontenc}
\usepackage{subcaption}
\usepackage{xcolor}
\usepackage{soul}
\usepackage{graphicx}
\usepackage{booktabs}
\usepackage{enumitem}
\usepackage{float}
\usepackage[style=numeric-comp,sorting=none,backend=biber,style=nature]{biblatex}
\newcommand\zy[1]{\textcolor{black}{#1}}
\newcommand\zyc[1]{\textcolor{black}{#1}} % erhanbas-checked
\newcommand{\ts}{\textsuperscript}

\usepackage[left=2.5cm,top=2.5cm,right=2.5cm,bottom=2.5cm,bindingoffset=1.0cm]{geometry}
\usepackage{authblk}
\title{Decipher-MR: A Vision-Language Foundation Model for 3D MRI Representations}

\date{}
\newcommand*\samethanks[1][\value{footnote}]{\footnotemark[#1]}
\author{Zhijian Yang\textsuperscript{1,*}, Noel DSouza\textsuperscript{1}, Istvan Megyeri\textsuperscript{2}, Xiaojian Xu\textsuperscript{1}, Amin Honarmandi Shandiz\textsuperscript{2}, Farzin Haddadpour\textsuperscript{1}, Krisztian Koos\textsuperscript{2}, Laszlo Rusko\textsuperscript{2}, Emanuele Valeriano\textsuperscript{1}, Bharadwaj Swaninathan\textsuperscript{1}, Lei Wu\textsuperscript{1}, Parminder Bhatia\textsuperscript{1}, Taha Kass-Hout\textsuperscript{1}, Erhan Bas\textsuperscript{1,*}\samethanks\\
\textsuperscript{1} GE Healthcare, Seattle, USA \\
\textsuperscript{2} GE Healthcare, Budapest, Hungary \\
* Corresponding author \\ Zhijian Yang; Erhan Bas: \{zhijian.yang;erhan.bas\}@gehealthcare.com
}
\begin{document}
\flushbottom
\maketitle
\vspace{-2em}
\thispagestyle{empty}

\maketitle

\begin{abstract}
Magnetic Resonance Imaging is a critical imaging modality in clinical diagnosis and research, yet its complexity and heterogeneity hinder scalable, generalizable machine learning. Although foundation models have revolutionized language and vision tasks, their application to MRI remains constrained by data scarcity and narrow anatomical focus. We present Decipher-MR, a 3D MRI-specific vision-language foundation model trained on 200,000 MRI series from over 22,000 studies spanning diverse anatomical regions, sequences, and pathologies. Decipher-MR integrates self-supervised vision learning with report-guided text supervision to build robust representations for broad applications. To enable efficient use, Decipher-MR supports a modular design that enables tuning of lightweight, task-specific decoders attached to a frozen pretrained encoder. Following this setting, we evaluate Decipher-MR across disease classification, demographic prediction, anatomical localization, and cross-modal retrieval, demonstrating consistent improvements over existing foundation models and task-specific approaches. These results position Decipher-MR as a versatile foundation for MRI-based AI in clinical and research settings.
\end{abstract}

\section{Introduction}
Magnetic Resonance Imaging (MRI) is a cornerstone of modern medical imaging, providing detailed, non-invasive visualization of soft tissues across various anatomical regions. It plays a crucial role in diagnosing and monitoring conditions in neurology, cardiology, musculoskeletal health, etc. However, the complexity of MR images, compounded by variations in acquisition protocols and sequences, presents significant challenges for automated analysis \cite{mri_challenge}. While traditional machine learning techniques have advanced image analysis, they still rely heavily on large labeled datasets and often struggle to generalize across different scanners, data sources, and clinical tasks, limiting their scalability and clinical reliability.

In response to these limitations, foundation models, large-scale deep learning models pre-trained on diverse and often unlabeled data, have emerged as a transformative approach in artificial intelligence. These models, which have achieved remarkable success in fields such as natural language processing (e.g., GPT)\cite{openai2024gpt4technicalreport} and computer vision (e.g., CLIP, DINO)\cite{clip}\cite{oquab2024dinov}\cite{MAE}, learn general-purpose representations that can be effectively adapted to various downstream tasks. By utilizing self-supervised learning and vast heterogeneous unlabeled data, foundation models help reduce the dependency on manual annotations and offer improved robustness and transferability across domains.

Recently, foundation models have gained increasing attention in medical imaging as a step toward building unified feature extractors for diverse medical tasks and application development. Extensive research has explored both vision-only and vision-language pretraining for specific modalities—such as X-ray\cite{raddino}\cite{raydino}, CT\cite{merlin}\cite{yang2024advancing}, and pathology images\cite{pathology_foundation1}\cite{pathology_foundation2}-demonstrating the effectiveness of both approaches in capturing modality-specific features, with the latter excelling at capturing cross-modal representations. Some foundation models are designed to leverage a combination of medical imaging modalities, providing a universal solution that is adaptable across various image types\cite{medimageinsight}\cite{medcoss}\cite{biomedclip}. These models support a broad spectrum of applications, including classification, segmentation, and retrieval, sometimes achieving remarkable few-shot or zero-shot performance with minimal fine-tuning. Furthermore, certain foundation models are specifically tailored to address critical tasks in radiological use cases, such as segmentation\cite{biomedparse}\cite{cox2024brainsegfounder3dfoundationmodels}, while also enabling efficient zero-shot performance through effective prompting\cite{medsam}\cite{sammed2d}. 

Despite these advancements, the application of foundation models to MRI remains largely underdeveloped. While some foundation models incorporate mixed modalities during pretraining — including MRI, CT, positron emission tomography, and microscopy — MRI data constitutes only a small fraction of the training set, leading to significant data imbalance. Additionally, existing MRI-specific foundation models often focus on narrow anatomical regions\cite{cox2024brainsegfounder3dfoundationmodels}\cite{brain_mri_foundation}\cite{wang2025triadvisionfoundationmodel}\cite{brainsegfounder} limiting their generalizability. Given MRI’s inherent heterogeneity, stemming from variations in scanner hardware, pulse sequences, clinical protocols, and its three-dimensional nature, there is a need for the development of large-scale, 3D MRI-specific foundation models trained on diverse datasets spanning different anatomical regions, imaging sequences, and disease conditions. 

Considering this need, we introduce Decipher-MR, a 3D foundation model specifically designed for MRI. Decipher-MR is trained on a large-scale dataset more than 200,000 3D MRI series from over 22,000 studies, covering a diverse range of ages, body regions, imaging sequences, and anatomical structures. With radiology reports available for most MRI studies, we further enhance Decipher-MR through text supervision on top of self-supervised vision learning, improving its robustness and enabling zero-shot search capabilities. Our pretraining approach combines unimodal learning with a vision–language alignment stage, allowing the model to learn rich, clinically relevant representations.

Given the diversity of medical AI tasks and the inefficiency of fine-tuning large models for each, Decipher-MR is designed to support a modular and reusable development pipeline. Task-specific decoders can be flexibly attached and tuned atop the frozen pretrained encoder, enabling efficient adaptation without requiring full model retraining. To evaluate this, we assess Decipher-MR across a wide range of tasks in a frozen encoder setting, including disease and demographic prediction, imaging attribute classification, organ and anomaly segmentation and localization, as well as cross-modal tasks like text-to-image and image-to-text retrievals. Our model consistently outperforms state-of-the-art foundation models and matches or exceeds specialized end-to-end training approaches with faster convergence. These results establish Decipher-MR as a scalable and versatile foundation for MRI-based AI in clinical and research applications.

\begin{figure}[!ht]
  \centering
  \includegraphics[width=1.0\linewidth]{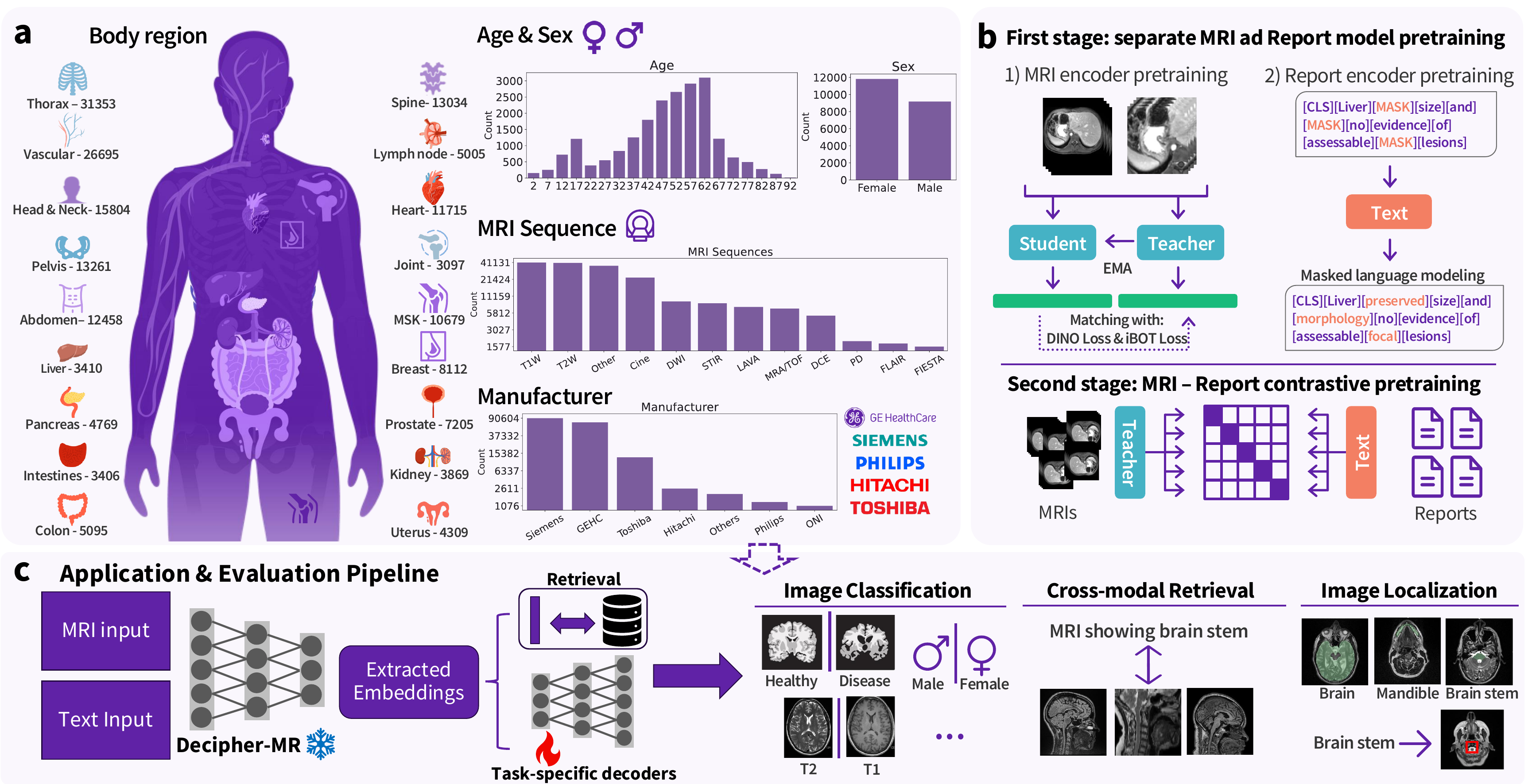}
    \caption{ Overview of Decipher-MR Dataset, Framework, and Evaluation (a) Distribution of the diverse pretraining dataset across age, sex, imaging sequences, body regions, and scanner manufacturers. The number of MRI series is shown for each body region, sequence type, and manufacturer, while the number of MRI studies is shown across age ranges and sex. (b) Overview of the two-stage pretraining framework of Decipher-MR. (c) Evaluation of Decipher-MR in a frozen encoder setup. Extracted embeddings are either used directly for retrieval tasks or paired with tunable and relatively lightweight decoders for specific tasks. Evaluation covers diverse tasks, including classification, cross-modal retrieval, and image localization.}
    \label{fig1}
\end{figure}

\section{Results}
\subsection{Decipher-MR Pretraining}
Decipher-MR was pretrained on a large and diverse MRI dataset comprising 22,594 studies and 203,233 series, with corresponding radiology reports available for 20,658 studies. The dataset spans a broad age range (0 to 90 years), covers a wide spectrum of body regions and MRI protocols, and features scans acquired from multiple vendors, including GE, Siemens, Philips, Toshiba, etc. \zyc{Detailed quantitative distributions are summarized in Fig.~\ref{fig1}a.} This diversity ensures robust representation across various imaging systems and clinical environments.

We employed a two-stage pretraining strategy (Fig.~\ref{fig1}b). In the first stage, we independently pretrained the image and text encoders using self-supervised learning: a student-teacher contrastive framework for the vision encoder and masked language modeling for the text encoder. In the second stage, we conducted joint image–report contrastive pretraining to align visual and textual representations. This approach enables the model to capture both global and fine-grained imaging features, while learning visual patterns aligned with key anatomical and pathological descriptors in the reports—thereby enhancing visual representation quality and enabling effective bidirectional retrieval between image scans and text descriptions. More details of pretraining can be found in Method 4.4.

 The Decipher-MR was evaluated across a diverse set of MRI-related tasks, including classification, retrieval, segmentation, localization, and visual grounding (Fig.~\ref{fig1}c). To address the challenge posed by increasing model sizes and the need for separate designs and training for each task, we aimed to establish a more efficient, modular development pipeline. This approach involves attaching and fine-tuning standardized, lightweight task-specific decoders to a pretrained but frozen encoder, thereby avoiding repeated tuning of large encoder networks (Fig.~\ref{fig1}c). In the following sections, we demonstrate and compare model performance across these tasks using this frozen encoder framework.

\begin{figure}[!ht]
  \centering
  \includegraphics[width=1.0\linewidth]{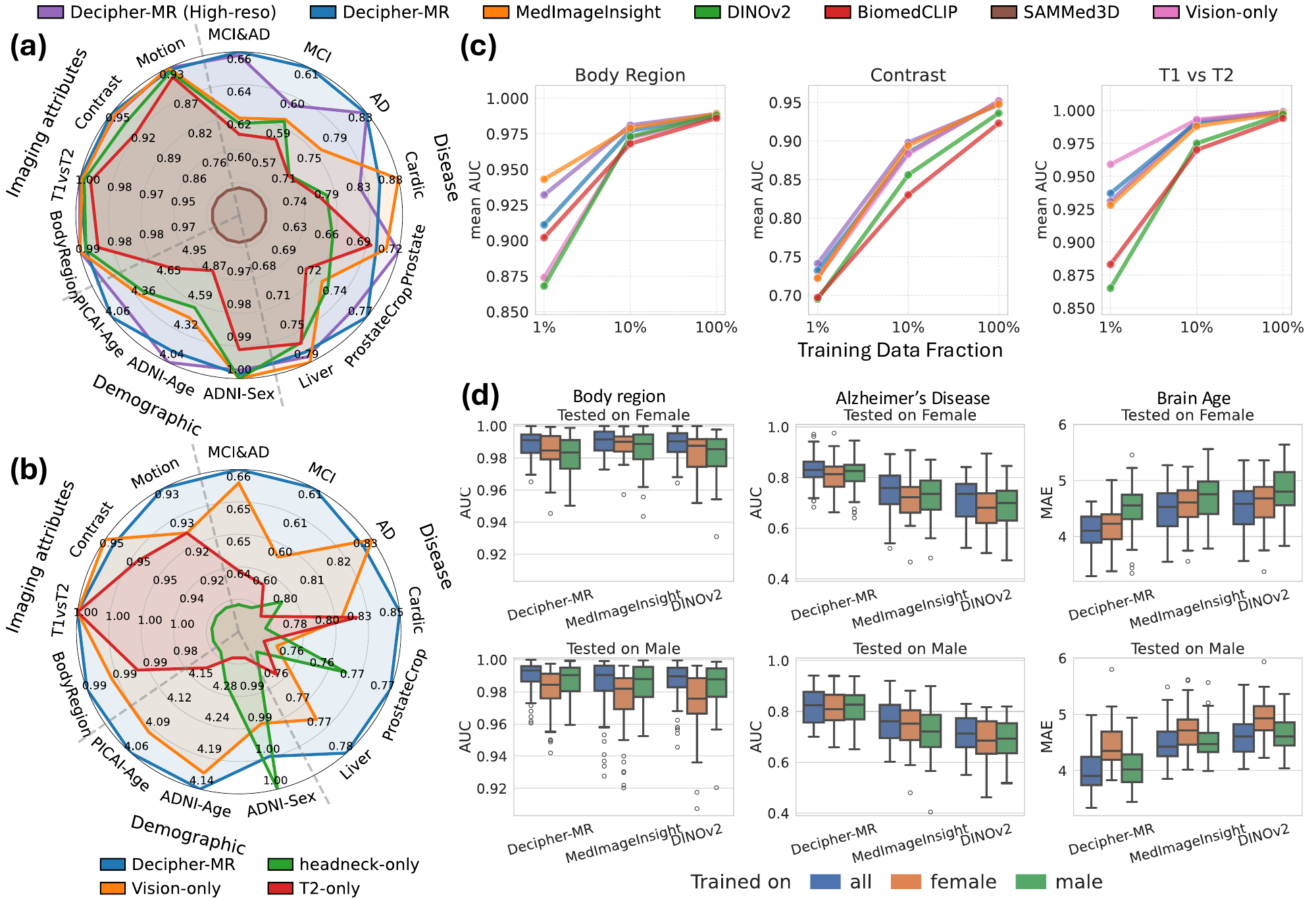}
    \caption{ Evaluation of Classification Probing Tasks
    (a) Comparison of foundation models on multiple medical image classification tasks using a simple MLP probe on CLS embeddings. 
    (b) Ablation of GE MR foundation models pretrained on different subsets of data: MRI-only, head and neck only, and T2-weighted only.
    (c) Performance under low-data regimes, where the MLP decoder is trained with varying proportions of labeled data. Full results across all tasks are in Supplementary Table \ref{tab:classification-disease}-\ref{tab:classification-imaging}. 
    (d) Bias analysis of the top three foundation models, comparing performance when training and testing within or across sexes. For all figures above, mean AUC was used as the evaluation metric, except for age prediction, which was assessed using MAE (lower is better).
    }
    \label{fig2}
\end{figure}

\subsection{Decipher-MR Learns Robust Visual Representation for Classification Tasks}
A broad range of clinical and radiological tasks involving MRI can be formulated as classification problems, including disease diagnosis, demographic prediction, body region identification, image sequence classification, and contrast detection. We benchmarked our method against state-of-the-art foundation models using a simple probing setup, where the pretrained encoder was frozen and only a lightweight three-layer MLP attached on top was fine-tuned across these varied tasks. Our model consistently outperformed DINOv2\cite{oquab2024dinov} and BiomedCLIP\cite{biomedclip} by a large margin, and demonstrated clear advantages over MedImageInsight\cite{medimageinsight}, achieving average gains of 2.9\% in disease classification (Supplementary Table~\ref{tab:classification-disease}), 3.0\% in demographic prediction (Supplementary Table~\ref{tab:classification-demo}), and 0.2\% in imaging-related attribute detection (Supplementary Table~\ref{tab:classification-imaging}; Fig.~\ref{fig2}a). \zyc{The statistical significance of differences across tasks is reported in Supplementary Table~\ref{tab:classification-pvalue}}. Performance differences between models became more pronounced under limited data settings (Supplementary Table~\ref{tab:classification-disease}-\ref{tab:classification-imaging}), especially in tasks such as imaging attribute detection where performance otherwise tends to saturate (Fig.~\ref{fig2}c). Further details on the classification tasks within each category, along with the datasets used, are provided in Methods Section 4.10 and Supplementary Table~\ref{tab:classification_task}.

We additionally performed bias analyses across sexes using the same probing setup (Fig.~\ref{fig2}d). For both body region and age prediction tasks, all three top-performing models performed better when trained and tested on the same sex, with further gains observed when using combined male and female training data. Nonetheless, Decipher-MR maintained superior performance even when tested across sexes (an average advantage of 5.5\% over MedimageInsight across three tasks), demonstrating its robustness to demographic variation.

Importantly, most of evaluations were performed without task-specific preprocessing (e.g., skull stripping, registration, or organ cropping) or fine-tuning, ensuring fair and direct model comparisons. Further performance improvements are possible through specialized model architectures and task-optimized data preprocessing. For instance, simply resizing MR images in-plane to match MedImageInsight’s input resolution substantially improved performance in tasks such as liver and prostate lesion classification, brain age regression, and attribute detection (Fig.~\ref{fig2}a). Additional cropping around the region of interest can further enhance performance for localized tasks—for instance, prostate lesion classification improved by 7\% when cropping around the prostate (ProstateCrop vs. Prostate in Fig.\ref{fig2}a and Supplementary Table \ref{tab:classification-disease}), further amplifying Decipher-MR’s advantage over other models. However, such cropping can impair performance on tasks that rely on broader anatomical context, such as age prediction from prostate MRI (Supplementary Table \ref{tab:classification-disease}), which benefits from information beyond the prostate region.
\\

\subsection{Decipher-MR Benefits from Text and Diversity in Pretraining Data}

To further assess the contributions of textual supervision and data diversity during pretraining, we conducted ablation studies using foundation models trained under varying conditions: vision-only (MRI without text), limited anatomical coverage (head and neck only), and restricted protocol types (T2-weighted only). 

Performance gains were observed with second-stage image–text contrastive pretraining, which outperformed models trained solely on imaging data (Fig.~\ref{fig2}b) on most tasks, \zyc{with significant gains in disease classification, brain age/sex prediction, and body region classification (Supplementary Table~\ref{tab:classification-pvalue})}). These results highlight the benefit of weak supervision from radiology reports and demonstrate the effectiveness of our two-stage pretraining strategy. Notable improvements were found in cardiac disease classification (+5.0\%), prostate lesion classification (+2.4\%), and several low-data scenarios, including brain age regression (+3.2\%), body region detection (+4.2\%), and contrast detection (+1.6\%), with the latter two particularly pronounced under low-data scenarios (Fig.~\ref{fig2}c).

Moreover, controlling for the contribution of text supervision and comparing against models trained on images only, pretraining on a diverse MRI dataset consistently improved performance across most tasks. Specifically, it outperformed models trained on constrained subsets, achieving average gains of 1.7\%, 1.6\%, and 1.6\% over the head-neck-only model, and 1.3\%, 2.1\%, and 1.3\% over the T2-only model for disease classification, demographic prediction, and imaging-related attribute detection, respectively (Fig.\ref{fig2}b and Supplementary Table~\ref{tab:classification-disease}–\ref{tab:classification-imaging}). Notably, improvements were observed even on head-neck-focused tasks such as Alzheimer's disease classification (3.8\%) and brain-age prediction (3.1\%) compared to the head-neck-only model, highlighting the benefit of pretraining on diverse, non-task-specific anatomical regions. These evidences underscore the critical importance of anatomical and protocol diversity for developing generalizable foundation models (Fig.~\ref{fig2}b).

\begin{figure}[!ht]
  \centering
  \includegraphics[width=1.0\linewidth]{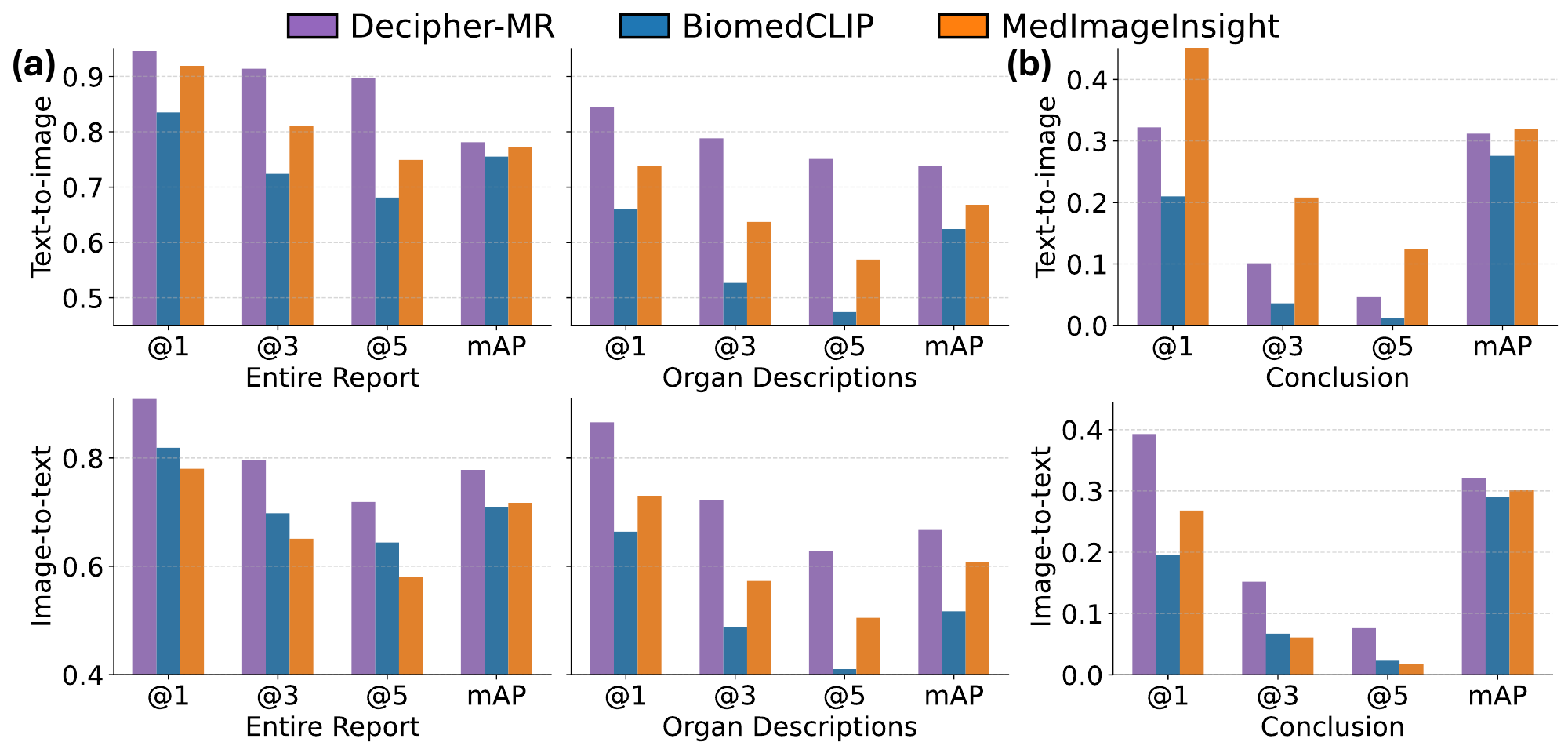}
    \caption{
Cross-Modal Retrieval Performance Across Two Datasets
(a) Performance on Source1 dataset: Body region retrieval
(b) Performance on Source2 dataset: Head–neck tumor pathology retrieval. Metrics include mean average precision (mAP) and precision@N—the proportion of queries where all top-N retrieved items match the correct category (i.e., body region or tumor sub-anatomical location).
}
    \label{fig3}
\end{figure}

\subsection{Decipher-MR Supports Cross-modality Retrieval}
Zero-shot cross-modal retrieval enables matching between MRI images and text—such as retrieving scans from text queries or linking relevant report descriptions to a given image—without the need for additional model fine-tuning. This capability is supported in our Decipher-MR through image–text contrastive pretraining. We evaluated its performance alongside two other medical vision–language models, MedImageInsight\cite{medimageinsight} and BioMedCLIP\cite{biomedclip}, across both in-domain and out-of-domain datasets.

On in-domain datasets, querying with either full reports or only the conclusion sections (Supplementary Fig.~\ref{supple_fig1}) retrieved the exact paired scan within the top 10 results in approximately 26\% of cases, \zyc{out of around 25{,}000 candidate MRI images from 2{,}500 studies}, substantially outperforming MedImageInsight (5.1\%) (Supplementary Table~\ref{tab:dp188_retrieval}).

For robust body region retrieval, Decipher-MR demonstrated consistently high success rates on the out-of-domain Source1 dataset (Fig. \ref{fig3}a). It achieved a top-3 localization success rate of 91.4\% using full reports and 78.8\% with short, single-organ descriptions (details in Supplementary Fig.~\ref{supple_fig2}b and Table~\ref{tab:retrieval_gt}), outperforming MedImageInsight’s 81.1\% and 63.7\%, respectively. Similar trends were observed in reverse (image-to-text) retrieval tasks. Beyond top-\emph{N} accuracy, Decipher-MR also achieved higher mean average precision (mAP), reflecting better overall ranking of relevant results. This improvement was particularly pronounced when using short and simple text queries which are more reflective of real-world applications where users are likely to use concise descriptions.

For tumor pathology retrieval in out-of-domain head and neck datasets using images as queries, Decipher-MR attained 39.3\% accuracy in retrieving conclusion sections describing similar tumor types within the same sub-anatomical region (details in Supplementary Fig.~\ref{supple_fig2}a and Table \ref{tab:retrieval_gt}), substantially outperforming BioMedCLIP (19.5\%) and MedImageInsight (26.8\%) (Fig. \ref{fig3}b). In the reverse text-to-image task, Decipher-MR (32.2\%) again outperformed BiomedCLIP (21.0\%) but lagged behind MedImageInsight (46.8\%). However, both models showed comparable mean average precision (mAP), with MedImageInsight’s higher top-N accuracy likely inflated because of lower retrieval diversity—it returned only 223 unique images (top three appeared 126 times), compared to 353 for Decipher-MR (top three appeared 36 times). This greater image-side diversity contributed to Decipher-MR’s superior performance in the reverse image-to-text retrieval.

\begin{figure}[!ht]
  \centering
  \includegraphics[width=1.0\linewidth]{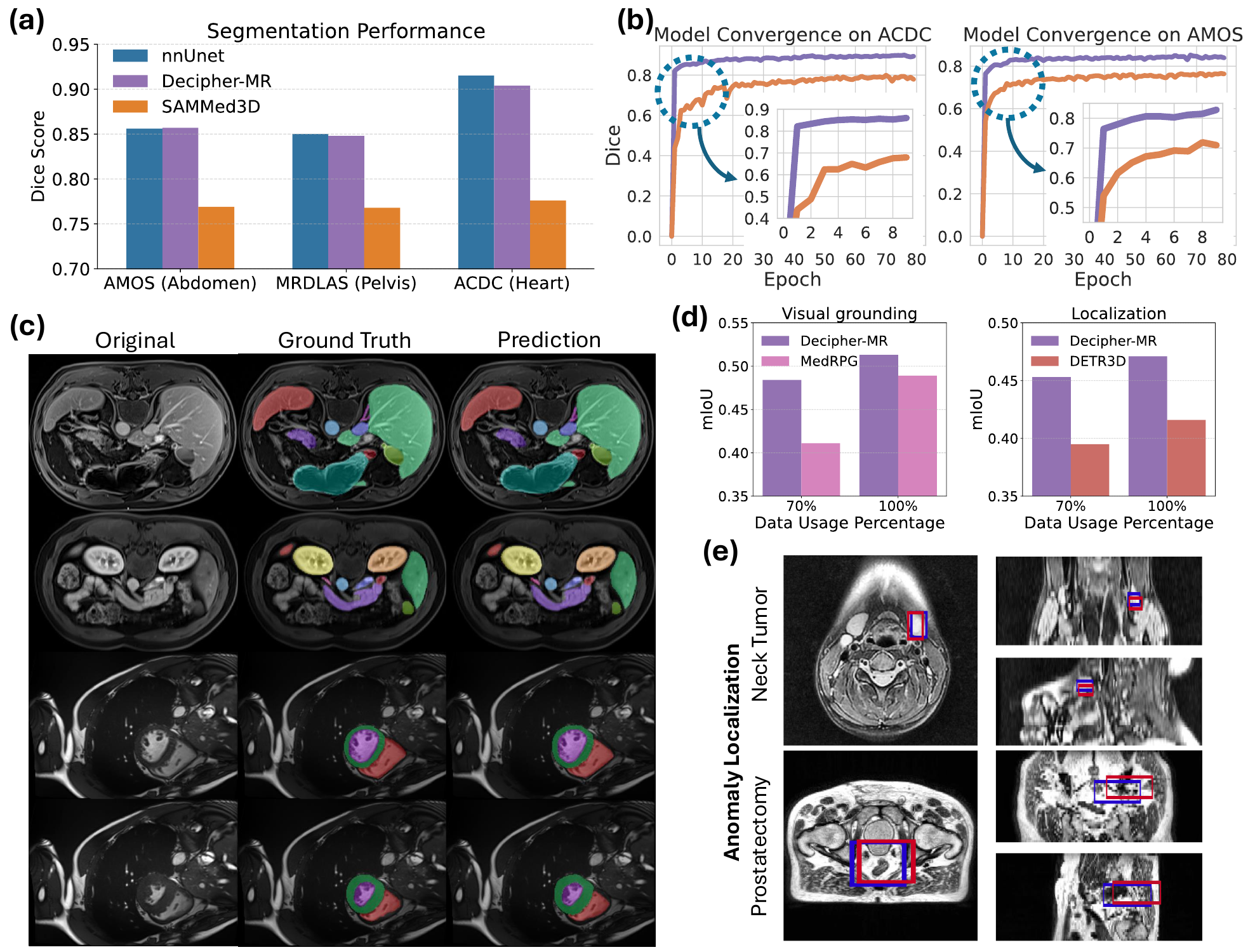}
     \caption{Evaluation of Segmentation and Localization Tasks
    (a) Anatomical segmentation performance across three body-region datasets, compared to nnUNet (end-to-end) and SAMed3D (frozen encoder).  
    (b) Validation Dice score over training epochs, comparing convergence speed between Decipher-MR and SAMed3D as frozen encoders.  
    (c) Qualitative examples of abdominal and cardiac segmentation using Decipher-MR as a frozen encoder.
    (d) Anomaly localization and visual grounding (e.g., missing organs, tumors), compared to end-to-end models.  
    (e) Qualitative examples of anomaly localization using Decipher-MR as a frozen encoder, including cases of neck tumors and prostatectomy. Ground truth bounding boxes are shown in blue, and predicted boxes in red.
    }

    \label{fig4}
\end{figure}

\subsection{Decipher-MR Enables Robust Medical Image Segmentation with Fast Convergence}

Three-dimensional anatomical segmentation plays a vital role in MRI-based clinical workflows, supporting diagnosis, surgical planning, and treatment monitoring. To evaluate the effectiveness of Decipher-MR for segmentation tasks, we tested it on three datasets covering abdominal (AMOS\cite{amos}), pelvic (MRDLAS), and cardiac regions (ACDC\cite{acdc}). A frozen encoder configuration was used to specifically assess the model’s feature extraction capabilities without fine-tuning encoder parameters.

Across all datasets, Decipher-MR paired with a ResNet-based decoder (Method 4.12) consistently outperformed SAMMed3D\cite{sammed3d} under the same frozen-encoder setting (Fig.~\ref{fig4}a). Remarkably, even though only the decoder was trained, the performance was comparable to nnUNet\cite{nnunet}, a widely used, fully tuned, state-of-the-art end-to-end segmentation framework in medical imaging (Fig.~\ref{fig4}a). Qualitative examples in Fig.~\ref{fig4}c show accurate segmentation of small abdominal organs and cardiac structures at both end-diastole (ED) and end-systole (ES), closely aligned with ground truth annotations. These results underscore the strong representational power of Decipher-MR across diverse anatomical regions.

Training convergence speed was another notable advantage. Decipher-MR converged rapidly, achieving a Dice score of 0.82 on ACDC and 0.77 on AMOS within just one epoch, which was markedly faster than SAMMed3D (Fig.~\ref{fig4}b). This highlights the value of strong pretrained initialization.

An interesting observation emerged regarding preprocessing resolution: segmentation performance with Decipher-MR consistently improved with higher-resolution resampling compared to the median voxel spacing commonly used in nnUNet. However, gains plateaued in datasets like AMOS, likely due to mixed image orientations, with performance saturating around the 25th percentile spacing (Supplementary Table~\ref{tab:segmentation-spacing}).

\subsection{Decipher-MR Supports Anomaly Localization and Visual Grounding}
Beyond classification and organ segmentation, accurate pathology or anomaly localization is critical for the clinical deployment of AI models, especially when integrating human input. To investigate this, we evaluated Decipher-MR on anomaly localization tasks involving head and neck tumors and surgically removed organs (e.g., prostatectomy, hepatectomy, and pneumonectomy) (Fig.~\ref{fig4}e), under two settings: standalone localization and visual grounding guided by human-provided text input describing anatomical location and pathology types. For each setting, we adopted state-of-the-art task-specific frameworks: DETR3D\cite{detr3d} for localization and MedRPG\cite{MedRPG} for visual grounding, replacing and initializing their visual encoders with Decipher-MR (Method 4.13).

In both scenarios, using Decipher-MR as a frozen encoder consistently outperformed the original end-to-end trained models, achieving a 14\% improvement in localization and an 11.5\% gain in visual grounding, as measured by mean Intersection over Union (mIoU; Fig.\ref{fig4}d) between predicted and groun-truth bounding boxes. These results underscore the strength of Decipher-MR’s visual representations for identifying and localizing different types of anomalies in MR images. Qualitative examples in Fig.\ref{fig4}e illustrate the localization performance on neck tumors and prostatectomy cases, showing good alignment between model predictions and ground truth annotations.

\section{Discussion}
In this study, we introduced Decipher-MR, an MRI-specific 3D vision-language foundation model pretrained on a large and diverse collection of MR datasets. We evaluated its capabilities across a comprehensive range of downstream tasks—including classification, cross-modal retrieval, organ segmentation, and anomaly localization—spanning multiple stages of the clinical workflow. Experimental results on both propriety and public datasets demonstrated that Decipher-MR is a powerful feature extractor for MR images, \zyc{with competitive inference speed (Supplementary Table 11)}. It consistently outperformed other foundation models on simple probing tasks and achieved performance on par with or exceeding state-of-the-art, task-specific models trained end-to-end. One particularly interesting finding is that Decipher-MR maintains strong performance even though its pretraining data primarily consists of individuals under the age of 70 and rarely include cases of Alzheimer’s disease (\zyc{six within 299 headneck or brain-scan patients aged $\geq$65 had reports referencing cognitive impairment or dementia.}) \cite{ad-age}. Despite this, it performs competitively on related tasks, suggesting a degree of generalization beyond the original pretraining distribution.

These advantages are largely attributed to the diversity of the pretraining data and the effectiveness of the pretraining strategy. Our ablation studies showed that including data from various body regions and MRI protocols led to improvements across most classification tasks, highlighting the robustness of the learned embeddings. This is consistent with observations that models pretrained on narrow or homogeneous data tend to underperform on body regions not represented during training \cite{wang2025triadvisionfoundationmodel}. A further benefit comes from the use of text supervision. Our two-stage pretraining pipeline allows flexible integration of both image-only data and image-text pairs. This not only enables cross-modal retrieval but also enhances classification performance, particularly in tasks related to body region, disease category, and demographic attributes.

Recognizing the critical importance of generalizability and bias mitigation in medical AI\cite{bias-healthai}, we conducted sex-based generalization evaluations across several probing tasks, including Alzheimer’s disease classification, brain age prediction, and body region recognition. Decipher-MR consistently outperformed other models when applied across sexes, demonstrating greater robustness in cross-sex scenarios. However, despite its advantage, Decipher-MR, like all foundation models, showed performance drops under cross-sex evaluation, highlighting the need for bias mitigation during fine-tuning\cite{dutt2023fairtune}\cite{bias-tuning}, though its more robust feature extraction could potentially offer a stronger foundation for fairer, more generalizable models.

As the field moves toward increasingly large and general-purpose models\cite{foundationai_medical}, a central challenge in medical AI is addressing the diverse range of clinical tasks without relying on bespoke architectures or repeated retraining. Decipher-MR tackles this by enabling modular reuse: a frozen, pretrained encoder paired with relatively lightweight, task-specific decoders—including MLP classifiers, ResNet-based segmentation heads, and localization/visual grounding modules. Despite minimal tuning, these components achieved performance comparable to or exceeding fully end-to-end models with rapid convergence, thus enabling efficient adaptation and streamlined development across a broad spectrum of medical imaging tasks. 

That said, task-specific data processing remains critical to unlocking the full potential of foundation models. We observed, for instance, that simple input resizing improved classification accuracy in certain tasks, while variations in resampling spacing significantly impacted segmentation outcomes. Local cropping proved beneficial for tasks targeting specific regions (e.g., tumor classification), but negatively affected those requiring broader anatomical context, such as age prediction. Additional pre-processing steps, such as skull-stripping \cite{DOSHI_skullstrip} and brain registration for neuroimaging tasks \cite{neural_data_processing}, or tumor localization and cropping before classification \cite{picai_localization_classification}, could further enhance performance.  Moreover, further decoder designs that jointly model classification and localization may offer performance boosts on more complex tasks.

Despite these advantages, several limitations remain. \zyc{First, we used Llama 3.3 to extract image-related information from radiology reports. Random human checks showed the task was straightforward and reliable, though full verification would further ensure consistency. Considering the human effort required, future validation could be performed using other LLMs as cross-checkers or judges.} Second, although Decipher-MR showed strong performance in cross-modal retrieval, particularly for anatomical region retrieval, its effectiveness in pathology-focused retrieval was more limited. This likely stems from insufficient diversity in textual report data during pretraining: \zyc{our reports came from a limited set of sources, where radiologists use highly standardized or repetitive terminology for similar conditions or negative findings, leading to many near-duplicated sentences and reduced generalization to more diverse texts. Increasing the diversity of pretraining data sources and incorporating LLM-based text augmentation to enrich report phrasing are potential directions to mitigate this.}  Third, we did not observe significant improvements in imaging attribute classification (e.g., sequence or contrast type) using additional text supervision. Incorporating DICOM metadata during pretraining may improve model capability in these areas and enable more radiology-relevant search functionalities. \zyc{Moreover, larger gains from the second-stage text–image alignment may require incorporating localization-aware supervision that aligns textual descriptions with specific regions, rather than treating the entire image holistically. Such region-level alignment could improve the model’s ability to localize anatomical structures and pathological findings.} Finally, while our frozen encoder setup has shown robust performance, encoder fine-tuning may be necessary to unlock further gains for certain applications. \zyc{With the idea of maintaining a universal backbone, we plan to explore tuning strategies like LoRA and adapters to balance efficiency and task-specific performance.}

In conclusion, we present Decipher-MR as one of the first multi-anatomy, multi-protocol MRI foundation models capable of supporting a wide range of downstream tasks. Our results demonstrate its potential to enable more efficient, modular, and scalable development of AI solutions for MRI data, paving the way for broader adoption of foundation models in clinical practice.

\section{Methods}
\subsection{Pretraining Dataset}
We compiled a diverse MRI dataset for model pretraining, encompassing various body regions such as the brain, head and neck, abdomen, pelvis, spine, knee, and others, with radiology reports available for the majority of the studies. To ensure data quality, we rigorously curated and preprocessed the scans, excluding those with missing slices, non-uniform spacing, low resolution, limited slice counts, or noisy voxel values. After preprocessing, the dataset consisted of 22,594 MRI studies and 203,233 MRI series, with reports for 20,658 studies. Each study includes a corresponding radiology report and several MRI series, covering different protocols such as T1, T2, DWI, and more. 85\% of studies were utilized in pretraining, with the remaining 15\% studies reserved for downstream evaluations and validations. 

\subsection{Evaluation Dataset}

We evaluated our Decipher-MR model using both proprietary and public datasets, benchmarking it against other methods.

\noindent Proprietary  datasets: We employed four proprietary datasets for different evaluation tasks:
\begin{itemize}[noitemsep, topsep=0pt, left=10pt]
    \item \zyc{Hold-out test set} of our pretraining datasets: Contains around 2,000 studies and 20,000 MRI series, with radiology reports for most studies. MRI sequence details, body regions, and contrast usage were extracted from DICOM headers.
    \item Source1 Dataset: Comprises 370 studies spanning various body regions, including the head, spine, abdomen, pelvis, extremities, and breast (Supplementary Fig.~\ref{supple_fig2}b and Table~\ref{tab:retrieval_gt}). A subset includes bounding box annotations for tumors and missing organs resulting from surgical removal.
    \item Source2 Head and Neck MRI Dataset: Includes 592 studies, each with multiple MRI series and reports. Most cases involve head and neck tumors, with annotations for tumor type and subregion (Supplementary Fig.~\ref{supple_fig2}a and Table~\ref{tab:retrieval_gt}).
    \item MRDLAS Segmentation Dataset: Provides segmentation masks for 10 pelvic organs across 93 pelvic T2 MRIs.
\end{itemize}

\noindent Public datasets: We also utilized seven public datasets for different evaluation tasks:
\begin{itemize}[noitemsep, topsep=0pt, left=10pt]
    \item ADNI: A dataset for Alzheimer's disease research, containing diagnoses for cognitively normal (CN), mild cognitive impairment (MCI), and Alzheimer's disease (AD) individuals, along with patient demographics. We used 1,732 baseline T1-weighted MRIs (\zyc{339 AD, 871 MCI, and 522 CN}) from ADNI1 and ADNI2 for classification evaluations.\cite{adni}
    \item PICAI: A prostate cancer dataset comprising 1,500 bi-parametric MRI (bpMRI) exams per patient, including T2-weighted (T2W), high b-value diffusion-weighted imaging (DWI), and apparent diffusion coefficient (ADC) maps, along with annotated prostate and tumor masks and diagnostic labels (benign (\zyc{1075 cases}) vs. malignant (\zyc{425 cases})) \cite{picai}.
    \item ACDC: A cardiac \zyc{cine-MRI} dataset with 150 exams across five subgroups, including four pathological conditions (myocardial infarction, dilated cardiomyopathy, hypertrophic cardiomyopathy, abnormal right ventricle) and one healthy group \zyc{with 30 cases per category}, along with segmentation masks for three heart regions \cite{acdc}. 
    \item LLD-MMRI: A multi-phase liver MRI dataset for lesion diagnosis, covering seven lesion types: three malignant (intrahepatic cholangiocarcinoma (\zyc{58 cases}), liver metastases (\zyc{51 cases}), hepatocellular carcinoma (\zyc{157 cases})) and four benign (hepatic hemangioma (\zyc{79 cases}), hepatic abscess (\zyc{54 cases}), hepatic cysts (\zyc{53 cases}), focal nodular hyperplasia (\zyc{46 cases})) \cite{lldmmri}.    
    \item MRART: A dataset comprising 438 T1-weighted structural MRI head images collected from 148 subjects, capturing three levels of motion artifacts: \zyc{level 1 (129 images), level 2 (109 images), and level 3 (198 images)} \cite{mrart}.    
    \item AMOS: An abdominal multi-organ segmentation dataset containing 60 MRI scans with voxel-level annotations for 15 organs \cite{amos}.
\end{itemize}

\subsection{Model Architecture}
For Decipher-MR, we employed a 3D Vision Transformer (ViT) \cite{vit} as the image encoder, using non-overlapping 8×8×8 volumetric patches. Each patch was flattened and passed through a linear projection to generate patch embeddings. Trilinear position embedding interpolation enables the model to handle inputs of variable sizes during inference and downstream applications. The rest of the architecture follows the standard ViT-Base configuration, consisting of 86 million parameters and a 768-dimensional embedding size; \zyc{and can be found in the original ViT paper \cite{vit}}. 

For the text encoder \cite{devlin-etal-2019-bert}, we adopted a BERT-based architecture initialized with PubMedBERT \cite{pubmedbert}, which was pre-trained on biomedical literature. It uses the same 768-dimensional embedding size and supports a maximum sequence length of 512 tokens. A linear projection was applied to the [CLS] token from both the image and text encoders to map them into a shared 512-dimensional multimodal embedding space.

\subsection{Model Pretraining}
Leveraging our extensive and diverse pretraining datasets, we implemented a two-stage pretraining strategy: (1) standalone pretraining of the image and text encoders using self-supervised learning methods, and (2) multimodal contrastive pretraining. This two-stage approach enables the model to capture both global and detailed imaging features, along with visual cues that reflect critical pathological and anatomical information described in the image reports. Additionally, multi-modal contrastive learning enhances cross-modal retrieval, allowing the model to search for relevant image scans from a database based on text input and vice versa.

For the first stage of image encoder pretraining, we employed DINOv2 \cite{oquab2024dinov}, a Student-Teacher Self-Supervised Learning strategy designed for vision transformers (ViTs). The student network learns to match the teacher network’s representations at both the image and patch levels. At the patch level, the student predicts the teacher’s features for masked regions using Masked Image Modeling (MIM). At the image level, a multi-crop contrastive objective aligns features from local crops with the teacher’s global representation. The teacher network processes full images and updates its parameters through an exponential moving average (EMA) of the student’s parameters.

For the first stage of text encoder pretraining, we fine-tuned a PubMedBERT model\cite{pubmedbert} on our medical report data using masked language modeling (MLM). In MLM, random tokens within the input text are masked, and the model learns to predict the original tokens based on the surrounding context. This approach enables the model to capture domain-specific linguistic patterns and improve its understanding of terminology in MRI reports.

In the second stage of pretraining, we leveraged the pretrained vision and text encoders from the first stage and further refined them using an Image-Report Contrastive Learning approach \cite{clip} to align image and text embeddings in a shared multi-modal embedding space. During training, MRI-report pairs are processed to generate their respective embeddings. The model encourages embeddings from matching image-report pairs to be close in the shared space while pushing apart embeddings from non-matching pairs. Details regarding the construction of image and text inputs, along with the data sampling strategy designed to enhance contrastive learning, are provided in Section 4.6.

\subsection{Data Augmentation}
To enable a generalist foundation model capable of handling inconsistencies in MRI resolution, size, and acquisition protocols across scanners and body regions, we tailored our vision encoder training to be robust to such inconsistencies. Beyond standard augmentations like rotation and flipping, we customized the cropping strategy used in DINO’s teacher network. DINOv2 includes two global crops, and we adapted them for MRI by defining one crop that covers the entire scan resized to the input dimensions, and another partial crop extracted at a randomly selected resolution (\zyc{0.5–1.5 mm in-plane, 1–6 mm out-of-plane}) to introduce scale diversity, allowing the same scan to be seen at multiple resolutions. From each of these two global crops, we generated local crops with the same resolutions for the student network, enabling the model to learn consistent image-level representations. In the second stage of multi-modal contrastive training, we preserved more complete contextual information while keeping robustness to resolution variation by randomly cropping within 75\%–100\% of the image size and resizing to a relatively larger input size to retain fine details and resolutions (Section 4.7). 

\subsection{Organ-Based Study-level Data Sampling}
Each MRI study includes a single report and multiple corresponding MRI series. First, to preserve study-level consistency, series and reports from the same study are never treated as non-matching pairs during training—that is, we avoid introducing false negative examples by ensuring that matching image-report pairs from the same study are always treated as positive pairs. More importantly, MRI data and reports describing different body regions or organs often exhibit clear distinctions, which can be easily separated in embedding space. However, variations in MRI data and reports from the same body region often reflect subtle but clinically important anatomical differences related to pathologies or other factors. To capture these nuances, we implement a batch-wise sampling strategy that, with high probability, selects MRI studies whose reports describe the same organ (i.e., “Mapped organ” in Supplementary Fig.~\ref{supple_fig1}) within each batch. We then use the organ-specific descriptions in the reports as the language input (i.e., “Details” in Supplementary Fig.~\ref{supple_fig1}). This approach enhances contrastive learning by focusing on differences within the same anatomical region, guiding the model to attend to localized features relevant to the targeted organ and thereby improving the extraction of meaningful representations and clinically significant information.

\subsection{Pretraining Data Processing}
MRI Processing: we first applied foreground cropping to all MRI images using Otsu's method \zyc{with 256 bins} to remove the black background. Each image was then rescaled to the range of -1 to 1 based on the 1st and 99th percentiles of pixel intensities. For the first-stage pretraining, we set the global crop size to 256×256×24 and the local crop size to 128×128×16, selected based on the median dimensions of our pretraining dataset. As described in Section~{4.5}, the first global crop was resized to match the target size while preserving the x-y aspect ratio, with black padding added as needed The second global crop was a random crop with variable resolution. \zyc{In the second-stage contrastive training, to capture a larger portion of the input image while minimizing information loss from resizing, we increased the crop size to 384×384×48.}
\\\\
Report Processing: we first used Llama 3.3 \cite{llama3herdmodels} to process the noisy MRI reports by removing study details unrelated to imaging and extracting organ-specific descriptions and conclusions into separate sections, including any identified abnormalities. We then mapped each organ section to a list of standard SNOMED anatomical entities \cite{snomed-ct} using AWS Comprehend Medical \cite{aws_comprehend_medical} for the sampling process described in Section 4.6. \zyc{We randomly inspected on average 30 processed reports per body region and refine the prompts to enhance robustness of the processing pipeline.} Additional details, including the prompt used and examples of reports before and after processing, are provided in Supplementary Fig.~\ref{supple_fig1}.

\subsection{Pretraining Implementation Details}
During the initial stage of vision encoder pretraining, the model was trained for 100 epochs with a batch size of 64 and a peak learning rate of 0.002. We employed a cosine learning rate scheduler with a linear warmup over the first 10 epochs, a teacher momentum consine schedule from 0.992 to 1, and a weight decay cosine schedule from 0.04 to 0.4, \zyc{and, for iBOT, a mask ratio randomly sampled in 0.1-0.5}. For the text encoder, we initialized it from PubMedBERT and fine-tuned it on our report data for 100 epochs with a learning rate of 5e-5 and an MLM \zyc{mask ratio of 0.15}. 

In the second stage—contrastive training—we used a batch size of 96 and trained for 20 epochs. A cosine learning rate scheduler with linear warmup over the first 2 epochs was employed, along with a weight decay schedule from 0.04 to 0.4. We assigned different peak learning rates to different components: 3e-4 for the linear projection head, and 3e-5 for both the vision and text encoders initialized from first stage. This learning rate configuration encourages effective adaptation to text supervision while preserving the pretrained knowledge from the first stage.

\subsection{Baselines}
We compared our models to different subsets of following baseline vision encoders, selected based on the nature of the tasks.
\begin{itemize}
    \setlength{\itemsep}{-0.5em} % Adjust this value to reduce space
  \item DinoV2\cite{oquab2024dinov}: A 2D vision model pretrained using a student-teacher self-supervised learning approach on large-scale natural image datasets.
  \item BiomedCLIP\cite{biomedclip}: A 2D vision-language model pretrained using a contrastive self-supervised learning approach on 15 million biomedical image-text pairs, sourced from 4.4 million biomedical articles in PubMed.
  \item SAMMed3D\cite{sammed3d}: A 3D medical segmentation model trained within the SAM framework using CT and MRI data. We specifically evaluated the model's 3D encoder as a feature extractor for various tasks, assuming it can effectively capture meaningful anatomical information from 3D medical images by leveraging its pre-training on segmentation tasks.
  \item MedImageInsight\cite{medimageinsight}: A vision-language model pretrained on medical images with associated text and labels from diverse modalities, including X-ray, CT, MRI, dermoscopy, OCT, fundus photography, ultrasound, histopathology, and mammography.
    \zyc{\item MRI-CORE\cite{mricore}: A 2D MRI-specific foundation model pretrained on more than 6 million slices from over 110{,}000 MRI volumes across 18 body locations.}
  \item \zyc{BrainSegFounder\cite{brainsegfounder}: A 3D brain MRI segmentation foundation model pretrained in a multi-stage self-supervised framework on large-scale brain MRI data from 41{,}400 participants. We use public Swin encoder as a feature extractor for our evaluation tasks.}
\end{itemize}

\subsection{Classification and Regression}
We evaluated Decipher-MR on MRI classification and regression tasks across five public datasets (ADNI, PICAI, AMOS, MRART, LLD-MMRI) as well as our proprietary dataset. These datasets cover a range of conditions and tasks including Alzheimer's disease, cardiac disease, prostate cancer, liver cancer, body region classification, image sequences, motion artifact detection, and demographic predictions (Supplementary Table \ref{tab:classification_task}).

For each task, we applied a 3-layer MLP with 768-dimensional hidden layers, 20\% dropout, and ReLU activations \zyc{(1.2M parameters)} to the CLS embedding of the backbone ViT, serving as the classification or regression head. All evaluations and comparisons followed a simple probing setup, with the weights of the pretrained vision backbone frozen. \zyc{For ACDC, we extracted embeddings from both the ED (end-diastolic) and ES ((end-systolic) phases of the cine-MRI and concatenated the resulting embeddings as input to the MLP.} \zyc{A consistent learning rate of 1e-4 with plateau decay was applied across all tasks, without intense hyper-parameter tuning. Classification performance and model comparisons were generally robust to the choice of learning rate, given the simplicity of the probing setup on the extracted embeddings. (Supplementary Table \ref{tab:classification-lr})} 

To accommodate different MRI image sizes and model input requirements, we randomly cropped subvolumes or selected slices from 3D MRI scans as input to the pretrained encoders during training. For Decipher-MR, we used images at their original resolution with an input size of (256, 256, 24) consistent with the first-stage pretraining. For other foundation models, input sizes were selected according to their specific requirements, with slice resizing applied for 2D methods following their respective preprocessing pipelines. During validation and testing, predictions from all subvolumes (obtained via a sliding window approach) or slices were averaged to produce a final prediction for each MRI image. 

We further assessed the impact of data preprocessing by evaluating performance on the PICAI dataset using MRI images cropped around the prostate based on provided masks. Additionally, we performed supplementary evaluations across all datasets using in-plane resizing to 512×512 for Decipher-MR, matching the higher-resolution input requirements of MedImageInsight. This variant is denoted as Decipher-MR (high-reso) in the figures.

To ensure reproducibility, we used 50 random train/validation/test splits for all experiments.  We report the average AUROC for classification tasks and the average mean absolute error (MAE) for age regression. \zyc{Additionally, we performed bias analyses by training and testing separately on male and female cohorts for brain age, Alzheimer’s disease (793 females; 993 males), and body-region prediction (2,630 females; 1,790 males) to evaluate cross-group generalizability.}

\subsection{Cross-Modal Retrieval}
We evaluated Decipher-MR’s zero-shot cross-modal retrieval performance on text-to-image and image-to-text tasks across three datasets, comparing it against two biomedical vision-language models: MedImageInsight and BiomedCLIP. 

Using the held-out test set from our pretraining dataset, we performed paired report-to-image retrieval, using both full radiology reports and conclusion sections as text queries. Retrieval performance was quantified as the proportion of cases in which the exact ground-truth MRI image (among approximately 25{,}000 MR images from 2{,}500 studies) appeared within the top 5, top 10, or top 20 retrieved results.

To test generalization, we evaluated on two out-of-domain datasets. On Source1 set, which contains reports and MRIs across nine body regions, we assessed \textbf{body region retrieval} using both full and organ-specific text sections. Performance was measured by the percentage of cases where all top N retrieved scans belonged to the same body region as text query, evaluated at top 1, 3 and 5 results. On the Source2 Head-Neck dataset, which includes cases with tumors across thirteen sub-anatomical regions, we performed \textbf{pathology-related retrieval} using conclusion sections. Detailed text queries and ground-truth categories are provided in Supplementary Fig.~\ref{supple_fig2} and Table~\ref{tab:retrieval_gt}. Performance was measured by the percentage of cases in which all top-N retrieved scans depicted a tumor in the same sub-region described by the query text. For both datasets, we also performed reverse image-to-text retrieval using MRI images as queries to retrieve the corresponding report sections. 

To ensure consistency with the multimodal pretraining setup and optimal alignment between visual and text embeddings, all MRI inputs were resized to 384×384×48 during retrieval evaluation for Decipher-MR. For BiomedCLIP and MedImageInsight, which use 2D inputs, the similarity between a 3D MRI and a text query was computed as the maximum cosine similarity across all slices.

\subsection{Segmentation}
We evaluated our model on anatomical organ segmentation tasks by fine-tuning a segmentation decoder while keeping the pretrained encoder’s weights frozen, using a five-fold cross-validation protocol. 

For the decoder, we adapted architectures from the SegResNet framework\cite{segresnet}, a ResNet-based segmentation model. The decoder was modified to leverage patch embeddings from the final and three intermediate layers of the encoder, using skip connections between encoder and decoder stages to preserve spatial detail and support segmentation of complex anatomical structures. \zyc{Overall, the resulting network contains approximately 7.6 million parameters.}

All images and masks were resampled to standardized voxel resolutions prior to training. In addition to resampling to median voxel spacing as in nnUNet, we also experimented with multiple resolution levels across each dataset—including the 25th percentile, 10th percentile, and smallest voxel spacing (i.e., highest resolution)—to assess the impact of input resolution on performance. \zyc{A learning rate of 1e-3 was used across all tasks. We generally found that a higher learning rate improved performance in all segmentation tasks, although excessively large values led to training instability and collapse.}

We benchmarked Decipher-MR against the 3D vision encoder from MedSAM3D, pretrained for segmentation, and the state-of-the-art nnUNet\cite{nnunet}, trained end-to-end as upper-bound references. Evaluation was conducted on three datasets: one propriety dataset, MRDLAS (pelvic organs), and two public datasets, AMOS and ACDC, targeting abdominal and cardiac organ segmentation, respectively. For MRDLAS and ACDC, we report the mean Dice score averaged over five cross-validation folds and all organs, using the same five-fold split as the nnUNet paper\cite{nnunet_revisited} for ACDC and directly adopting their reported score as the nnUNet performance. For AMOS, we used the official train/validation split\cite{amos}, reported the mean Dice across all organs on the validation set, and referenced the nnUNet performance from the original AMOS paper\cite{amos}.

\subsection{Localization and Visual Grounding}
We evaluated Decipher-MR on anomaly localization tasks under two settings: (1) visual-only localization, where bounding boxes are predicted from image input alone, and (2) visual grounding, where text prompts guide spatial prediction.
We adopted two representative frameworks: a transformer-based 3D localization model adapted from DETR3D~\cite{detr3d}, and the visual grounding model MedRPG~\cite{MedRPG}, for which we converted the original 2D ResNet backbone to 3D. In the latter, visual patch embeddings from a frozen vision encoder were fused with text embeddings via a multimodal transformer, and a shared regression token was used to predict 3D bounding boxes through a lightweight MLP head. A combination of L1 loss and generalized IoU was used for training.

MRI volumes were resampled to a standardized size of 256×256×40 voxels. A sliding window strategy was used during inference to accommodate GPU memory constraints. For grounding, synthetic prompts—referring to anatomical regions, sides (e.g., “right kidney”), and pathology types (e.g., “tumor”, “removal”)—were used during both training and evaluation to mimic real-world clinical language.

Experiments were conducted on subsets of the Source1 and MRDLAS datasets, which include bounding box annotations for conditions such as prostatectomy, brain tumors, hepatectomy, neck tumors, and pneumonectomy, covering head, neck, abdominal, and pelvic regions. For both tasks, we replaced the visual encoder in each framework with Decipher-MR (frozen) and compared against the original end-to-end trained models. We report mean Intersection over Union (mIoU), averaged over five independent runs with different data splits.

\section{Acknowledgment}
This study received no funding. 
\section{Author Contributions}
ZY designed the pretraining model with contributions form XX, NDS and IM on model design and data preparation. 
NDS, IM, XX, AH, FH, KK, LR worked on decoder design, carried out finetuning experiments.
EV, BS, LW, PB, TKH assisted experimental setup for finetuning and analysis of results. 
EB designed and directed the project with contributions from PB and TKH.
ZY wrote the majority of manuscript with inputs from NDS, IM, XX, AH, FH, KK, LR, and EB. 
All authors discussed the results and contributed to the final manuscript.
\section{Competing Interests}
\zyc{All authors are employees of GEHC. The authors declare no other
competing interests.}

\printbibliography
\clearpage  % Start new page
\setcounter{figure}{0}
\setcounter{table}{0}
% Change layout here
\newgeometry{left=0.8cm,top=2.5cm,right=0.8cm,bottom=2.5cm,bindingoffset=0.5cm}
\setlength{\tabcolsep}{4pt}
\renewcommand{\tablename}{Supplementary Table}
\renewcommand{\figurename}{Supplementary Figure}

\section{Supplementary}
\begin{figure}[!ht]
  \centering
  \includegraphics[width=1.0\linewidth]{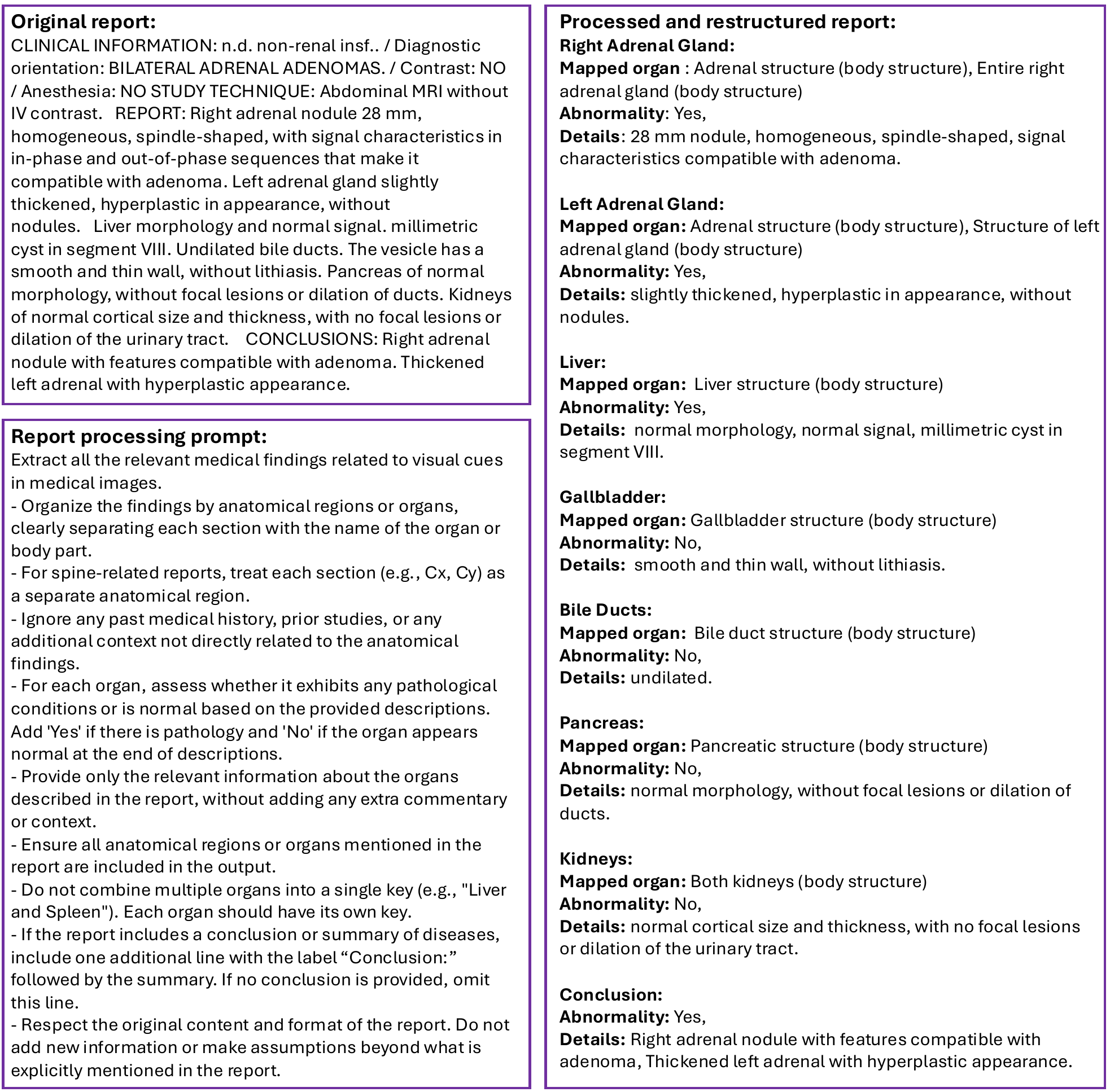}
    \caption{Details of the report processing procedure, including the prompt used for the LLM, along with an example of an original unprocessed report and the final processed and restructured output used for pretraining.}
    \label{supple_fig1}
\end{figure}
\begin{figure}[!ht]
  \centering
  \includegraphics[width=1.0\linewidth]{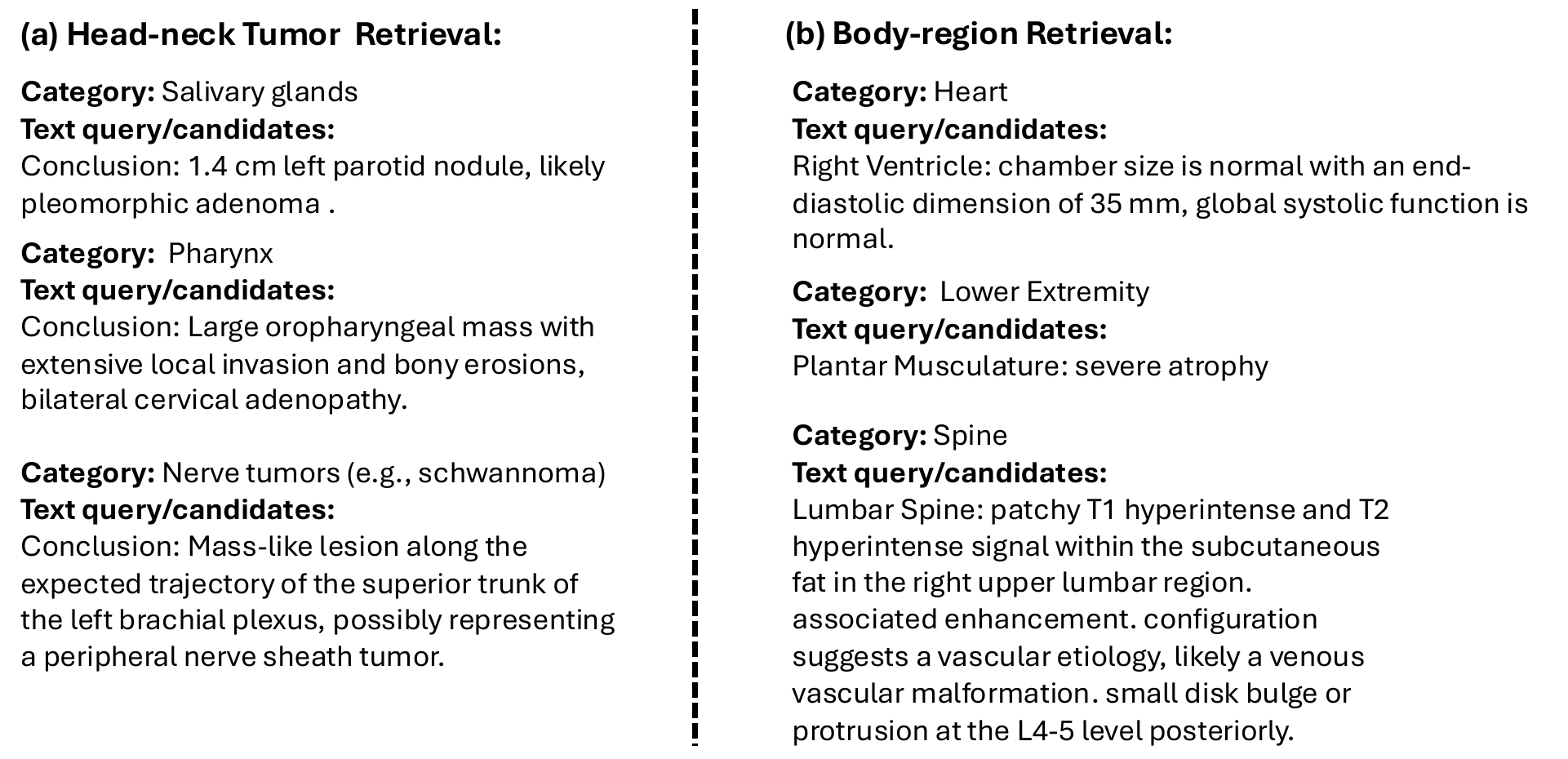}
    \caption{Examples of ground truth and text query/candidates in cross-modal retrieval tasks. (a)Head-neck tumor pathology retrieval (Source2 dataset): The ground-truth labels correspond to tumor types or specific anatomical subregions affected, and the text queries/candidates are the conclusion sections of reports describing the tumor. (b) Body region retrieval (Source1 dataset): The ground truth category corresponds to the anatomical body region. Text queries/candidates are derived from individual organ-specific descriptions in the report. Full-report queries/candidates are formed by concatenating all such descriptions from a single report.}
    \label{supple_fig2}
\end{figure}

\begin{table}[H]
    \centering
    \begin{tabular}{ccc}
    \hline
    & Body Region Categories (Source1) & Tumor Subregion Categories (Source2) \\
    \hline
    & Head & Gliomas \\
    & Spine & Meninges \\
    & Chest & Nerve tumors (e.g., schwannoma) \\
    & Heart & Pituitary \\
    & Abdomen & Skull base \\
    & Pelvis & Extra-axial metastasis \\
    & Breast & Salivary glands \\
    & Upper Extremity & Larynx \\
    & Lower Extremity & Pharynx \\
    & & Nasal cavity \\
    & & Oral cavity \\
    & & Orbit \\
    & & Neck (including thyroid) \\
    \end{tabular}
    \caption{Ground-truth category lists for the two cross-modal retrieval tasks. The first column lists nine body region categories used for body region retrieval with the Source1 dataset. The second lists thirteen tumor subregion categories from the Source2 dataset, representing tumor types or affected anatomical regions.}
    \label{tab:retrieval_gt}
\end{table}

\begin{table}[H]
    \centering
    \renewcommand{\arraystretch}{1.3}
    \begin{tabular}{p{3cm} p{13cm}}
    \hline
    \textbf{Dataset} & \textbf{Classification Tasks Performed} \\
    \hline
    ADNI & Multi-class classification (CN vs. MCI vs. AD); binary classification (CN vs. MCI, CN vs. AD); brain age prediction (regression); and sex classification. \\
    ACDC & Classification of five cardiac disease subtypes. \\
    PICAI & Benign vs. malignant prostate tumor classification; age prediction from prostate MRI. \\
    LLD-MMRI & Classification of seven liver lesion types (four malignant, three benign). \\
    MRART & Classification of three levels of motion artifacts. \\
    Proprietary Dataset & Classification of 11 body regions, T1 vs. T2 image sequences, and binary contrast presence, based on labels extracted from DICOM headers. \\
    \hline
    \end{tabular}
    \caption{Classification tasks conducted on each dataset.}
    \label{tab:classification_task}
\end{table}

\begin{table}[H]
    \centering
    \begin{tabular}{l ccccc}
    \hline
 & \multicolumn{5}{c}{ADNI}\\
         Models&  10\% (138) &  25\% (346) &  all (1384)& MCI (1114) & AD (688)\\
         \hline
         Decipher-MR (high-reso)
&  \bf{0.612$\pm$0.038}
&  \bf{0.631$\pm$0.027}
&  0.655$\pm$0.029
&  0.596$\pm$0.046
&  0.827$\pm$0.043

\\
         \hline
         Decipher-MR
&  0.603$\pm$0.037
&  0.622$\pm$0.029
&  \bf{0.657$\pm$0.028}
&  \bf{0.615$\pm$0.037}
&  0.826$\pm$0.046

\\
         Decipher-MR (1\ts{st} stage)
&  0.596$\pm$0.038
&  0.62$\pm$0.028
&  0.655$\pm$0.026
&  0.600$\pm$0.040
&  \bf{0.829$\pm$0.044}
\\
         MedImageInsight
&  0.561$\pm$0.036
&  0.585$\pm$0.034
&  0.62$\pm$0.026
&  0.589$\pm$0.043
&  0.757$\pm$0.048

\\
         DINOv2
&  0.557$\pm$0.036
&  0.582$\pm$0.032
&  0.617$\pm$0.028
&  0.586$\pm$0.043
&  0.717$\pm$0.053

\\
         BiomedCLIP
&  0.552$\pm$0.034
&  0.567$\pm$0.031
&  0.611$\pm$0.032
&  0.579$\pm$0.045
&  0.707$\pm$0.054

\\
         SAMMed-3D
&  0.540$\pm$0.042
&  0.553$\pm$0.037
&  0.581$\pm$0.035
&  0.553$\pm$0.050
&  0.664$\pm$0.059
\\
         MRI-CORE
&  0.540$\pm$0.035
&  0.544$\pm$0.028
&  0.562$\pm$0.030
&  0.553$\pm$0.043
&  0.655$\pm$0.056
 
\\
         BrainSegFounder
&  0.509$\pm$0.031
&  0.518$\pm$0.033
&  0.528$\pm$0.036
&  0.519$\pm$0.053
&  0.541$\pm$0.060
\\
\hline
         Decipher-MR Headneck
&  0.581$\pm$0.042
&  0.596$\pm$0.031
&  0.637$\pm$0.030
&  0.589$\pm$0.043
&  0.799$\pm$0.047

\\
         Decipher-MR T2
&  0.575$\pm$0.036
&  0.606$\pm$0.029
&  0.642$\pm$0.029
&  0.594$\pm$0.040
&  0.792$\pm$0.040

\\
         \hline
    \end{tabular}
        \begin{tabular}{l ccc ccc}
    \hline
 & \multicolumn{1}{c}{ACDC} & \multicolumn{3}{c}{LLD-MMRI}\\
         Models&   all (120)&   10\% (39) & 25\% (99) & all (398)\\
         \hline
         Decipher-MR (high-reso)
&  0.822$\pm$0.062
&  0.675$\pm$0.049
&  0.717$\pm$0.047
&  0.781$\pm$0.043

\\
         \hline
         Decipher-MR

&  0.850$\pm$0.063
&  0.672$\pm$0.053
&  0.714$\pm$0.050
&  0.775$\pm$0.036

\\
         Decipher-MR (1\ts{st} stage)
&  0.809$\pm$0.070
&  0.658$\pm$0.052
&  0.707$\pm$0.047
&  0.770$\pm$0.038
\\
         MedImageInsight
&  \bf{0.876$\pm$0.063}
&  \bf{0.676$\pm$0.046}
& \bf{0.729$\pm$0.047}
& \bf{0.788$\pm$0.037}

\\
         DINOv2
&  0.776$\pm$0.084
&  0.670$\pm$0.052
&  0.713$\pm$0.047
&  0.763$\pm$0.042

\\
         BiomedCLIP
&  0.766$\pm$0.069
&  0.658$\pm$0.055
&  0.702$\pm$0.051
&  0.764$\pm$0.042

\\
         SAMMed-3D
&  0.691$\pm$0.084
&  0.580$\pm$0.043
&  0.596$\pm$0.037
&  0.632$\pm$0.048

\\
         MRI-CORE
&  0.688$\pm$0.081
&  0.614$\pm$0.055
&  0.644$\pm$0.046
&  0.690$\pm$0.042

\\
         BrainSegFounder 
&  0.671$\pm$0.072
&  0.562$\pm$0.044
&  0.577$\pm$0.038
&  0.605$\pm$0.043

\\
\hline
         Decipher-MR Headneck
&  0.756$\pm$0.076
&  0.651$\pm$0.057
& 0.704$\pm$0.057
& 0.757$\pm$0.043
\\
         Decipher-MR T2
&  0.820$\pm$0.074
&  0.651$\pm$0.051
&  0.695$\pm$0.050
&  0.761$\pm$0.035

\\
         \hline
    \end{tabular}
        \begin{tabular}{l ccc ccc}
    \hline
 & \multicolumn{3}{c}{PICAI}& \multicolumn{3}{c}{PICAI-Crop}\\
         Models& 10\% (120) & 25\% (300) & all (1200)&  10\% (120) & 25\% (300) & all (1200)\\
         \hline
         Decipher-MR (high-reso)
&  \bf{0.632$\pm$0.053}
&  \bf{0.671$\pm$0.050}
&  \bf{0.720$\pm$0.044}
&  0.688$\pm$0.044
&  0.717$\pm$0.039
&  0.758$\pm$0.044

\\
         \hline
         Decipher-MR

&  0.620$\pm$0.060
&  0.660$\pm$0.053
&  0.700$\pm$0.040
&  \bf{0.691$\pm$0.044}
&  \bf{0.722$\pm$0.036}
&  \bf{0.770$\pm$0.039}

\\
         Decipher-MR (1\ts{st} stage)
&  0.616$\pm$0.059
&  0.658$\pm$0.055
&  0.696$\pm$0.046
&  0.686$\pm$0.047
&  0.717$\pm$0.038
&  0.752$\pm$0.042
\\
         MedImageInsight
&  0.617$\pm$0.055
&  0.657$\pm$0.054
&  0.708$\pm$0.041
&  0.676$\pm$0.044
&  0.696$\pm$0.045
&  0.723$\pm$0.045

\\
         DINOv2
&  0.569$\pm$0.051
&  0.599$\pm$0.058
&  0.660$\pm$0.037
&  0.684$\pm$0.045
&  0.702$\pm$0.048
&  0.731$\pm$0.040

\\
         BiomedCLIP
&  0.594$\pm$0.049
&  0.631$\pm$0.048
&  0.695$\pm$0.038
&  0.666$\pm$0.042
&  0.687$\pm$0.041
&  0.709$\pm$0.044

\\
         SAMMed-3D
&  0.525$\pm$0.056
&  0.556$\pm$0.054
&  0.599$\pm$0.048
&  0.648$\pm$0.056
&  0.654$\pm$0.050
&  0.663$\pm$0.050
\\
         MRI-CORE
&  0.523$\pm$0.051
&  0.539$\pm$0.054
&  0.576$\pm$0.046
&  0.630$\pm$0.042
&  0.653$\pm$0.042
&  0.678$\pm$0.048

\\
         BrainSegFounder
&  0.511$\pm$0.055
&  0.513$\pm$0.061
&  0.560$\pm$0.052
&  0.651$\pm$0.045
&  0.655$\pm$0.043
&  0.670$\pm$0.038

\\
\hline
         Decipher-MR Headneck
&  0.622$\pm$0.069
&  0.659$\pm$0.057
&  0.703$\pm$0.047
&  0.698$\pm$0.041
&  0.727$\pm$0.036
&  0.763$\pm$0.036
\\
         Decipher-MR T2
&  0.617$\pm$0.065
&  0.658$\pm$0.053
&  0.699$\pm$0.047
&  0.678$\pm$0.053
&  0.716$\pm$0.035
&  0.750$\pm$0.041

\\
         \hline
    \end{tabular}
    \caption{Performance on disease classification tasks. \zy{Mean AUC and standard deviation across 50 random data splits are reported for all models and tasks. Results are shown at different low-data percentages, where only a subset of the training set is used to fine-tune the model, with the number of subjects for each subset indicated next to the percentage.}}
    \label{tab:classification-disease}
\end{table}

\begin{table}[H]
    \centering
    \begin{tabular}{l cccccc ccc ccc}
    \hline
 & \multicolumn{3}{c}{PICAI}& \multicolumn{3}{c}{PICAI-Crop} \\
         &  \multicolumn{3}{c}{Age ($\downarrow$)}&  \multicolumn{3}{c}{Age ($\downarrow$)}\\
         Models& 10\% (120) & 25\% (300) & all (1200)&  10\% (120) & 25\% (300) & all (1200)\\
         \hline
         Decipher-MR (high-reso)
&  4.774$\pm$0.333
&  4.528$\pm$0.281
&  4.214$\pm$0.284
&  5.296$\pm$0.324
&  5.103$\pm$0.341
&  4.802$\pm$0.331

\\         \hline
         Decipher-MR
& \bf{4.679$\pm$0.300}
&  4.435$\pm$0.285
&  \bf{4.064$\pm$0.264}
& 5.220$\pm$0.334
&  4.971$\pm$0.324
&  4.676$\pm$0.285

\\
         Decipher-MR (1\ts{st} stage)
& 4.689$\pm$0.295
&  \bf{4.406$\pm$0.281}
&  4.087$\pm$0.258
&  \bf{5.150$\pm$0.320}
&  \bf{4.863$\pm$0.287}
&  \bf{4.532$\pm$0.271}
\\
         MedImageInsight
&  4.899$\pm$0.341
&  4.663$\pm$0.304
&  4.386$\pm$0.306
&  5.367$\pm$0.335
&  5.239$\pm$0.340
&  5.126$\pm$0.324
\\
         DINOv2
&  4.940$\pm$0.328
& 4.722$\pm$0.306
& 4.435$\pm$0.296
&  5.347$\pm$0.334
& 5.216$\pm$0.323
& 5.066$\pm$0.321

\\
         BiomedCLIP
&  5.114$\pm$0.350
& 4.958$\pm$0.323
&4.772$\pm$0.316
&  5.444$\pm$0.355
& 5.323$\pm$0.343
&5.206$\pm$0.326

\\
         SAMMed-3D
&  5.520$\pm$0.344
& 5.436$\pm$0.346
&5.295$\pm$0.344
&  5.574$\pm$0.364
& 5.553$\pm$0.349
&5.534$\pm$0.334
\\
         MRI-CORE
&  5.279$\pm$0.347
& 5.142$\pm$0.333
&4.942$\pm$0.337
&  5.609$\pm$0.365
& 5.576$\pm$0.365
&5.527$\pm$0.339

\\
         BrainSegFounder
&  5.604$\pm$0.360
& 5.596$\pm$0.354
&5.577$\pm$0.356
&  5.667$\pm$0.367
& 5.664$\pm$0.359
&5.648$\pm$0.355

\\
\hline
         Decipher-MR Headneck
&  4.716$\pm$0.312
&  4.488$\pm$0.291
&  4.178$\pm$0.265
&  5.197$\pm$0.334
&  4.950$\pm$0.322
&  4.609$\pm$0.283
\\
         Decipher-MR T2
&  4.700$\pm$0.325
& 4.378$\pm$0.258
& 4.160$\pm$0.236
&  5.195$\pm$0.323
& 4.914$\pm$0.297
& 4.595$\pm$0.278

\\

         \hline
    \end{tabular}
        \begin{tabular}{l cccccc}
    \hline
 & \multicolumn{6}{c}{ADNI}\\
         &  \multicolumn{3}{c}{Sex ($\uparrow$)}&  \multicolumn{3}{c}{Age ($\downarrow$)}\\
         Models&  10\% (138) & 25\% (346) & all (1384) &  10\% (138) & 25\% (346) & all (1384)\\
         \hline
         Decipher-MR (high-reso)
&  0.979$\pm$0.008
&  0.989$\pm$0.005
&  0.994$\pm$0.004
&  \bf{4.523$\pm$0.339}
&  \bf{4.294$\pm$0.316}
&  \bf{4.044$\pm$0.262}

\\         \hline
         Decipher-MR
&  \bf{0.983$\pm$0.007}
&  \bf{0.992$\pm$0.004}
&  0.995$\pm$0.004
&  4.559$\pm$0.324
&  4.353$\pm$0.302
&  4.143$\pm$0.264

\\
         Decipher-MR (1\ts{st} stage)
&  0.978$\pm$0.007
&  0.989$\pm$0.006
&  0.994$\pm$0.004
&  4.708$\pm$0.317
&  4.437$\pm$0.293
&  4.167$\pm$0.246
\\
         MedImageInsight
&  \bf{0.983$\pm$0.009}
&  \bf{0.992$\pm$0.005}
&  \bf{0.996$\pm$0.004}
&  4.883$\pm$0.314
&  4.704$\pm$0.280
&  4.460$\pm$0.288
\\
         DINOv2
&  0.978$\pm$0.010
&  \bf{0.992$\pm$0.005}
&  \bf{0.996$\pm$0.004}
& 5.026$\pm$0.288
&  4.817$\pm$0.289
&  4.556$\pm$0.278

\\
         BiomedCLIP
&  0.963$\pm$0.016
&  0.980$\pm$0.009
&  0.989$\pm$0.006
& 5.225$\pm$0.288
&  5.095$\pm$0.305
&  4.902$\pm$0.290

\\
         SAMMed-3D
&  0.944$\pm$0.022
&  0.956$\pm$0.019
&  0.963$\pm$0.013
& 5.390$\pm$0.339
&  5.266$\pm$0.314
&  5.162$\pm$0.306

\\
         MRI-CORE
&  0.957$\pm$0.016
&  0.973$\pm$0.009
&  0.983$\pm$0.007
& 5.426$\pm$0.310
&  5.316$\pm$0.297
&  5.191$\pm$0.292

\\
         BrainSegFounder
&  0.626$\pm$0.052
&  0.671$\pm$0.070
&  0.774$\pm$0.057
& 5.777$\pm$0.336
&  5.772$\pm$0.334
&  5.773$\pm$0.336

\\
\hline
         Decipher-MR Headneck
&  0.987$\pm$0.006
&  0.993$\pm$0.005
&  0.996$\pm$0.004
&  4.889$\pm$0.314
&  4.609$\pm$0.272
&  4.301$\pm$0.244
\\
         Decipher-MR T2
&  0.966$\pm$0.013
&  0.983$\pm$0.007
&  0.992$\pm$0.004
& 4.948$\pm$0.296
&  4.628$\pm$0.293
&  4.339$\pm$0.263

\\

         \hline
    \end{tabular}
    \caption{Performance on demographic predictions. \zy{Mean AUC (for sex classification) or MAE (for age regression), along with the standard deviation across 50 random data splits, are reported for all models and tasks. Results are shown at different low-data percentages, where only a subset of the training set is used to fine-tune the model, with the number of subjects for each subset indicated next to the percentage.}}
    \label{tab:classification-demo}
\end{table}

\begin{table}[H]
    \centering
        \begin{tabular}{l cccccccccccc}
    \hline
         &  \multicolumn{3}{c}{Body region}&  \multicolumn{3}{c}{Image sequence}\\
         Models&  1\% (35) & 10\% (353) & all (3635) &  1\% (110) & 10\% (1100) & all (11008) \\
         \hline
          Decipher-MR (high-reso)
&  0.932$\pm$0.021
&  \bf{0.981$\pm$0.006}
&  \bf{0.989$\pm$0.006}
&  0.931$\pm$0.019
&  \bf{0.992$\pm$0.003}
&  \bf{0.999$\pm$0.001}
\\         \hline
          Decipher-MR
&  0.911$\pm$0.028
&  0.977$\pm$0.008
&  0.988$\pm$0.005
&  0.937$\pm$0.018
&  0.991$\pm$0.003
&  \bf{0.999$\pm$0.001}
\\
         Decipher-MR (1st stage)
&  0.874$\pm$0.028
&  0.972$\pm$0.009
&  0.987$\pm$0.005
&  \bf{0.959$\pm$0.014}
&  0.991$\pm$0.003
&  \bf{0.999$\pm$0.001}
\\
        MedImageInsight
&  \bf{0.943$\pm$0.022}
&  0.979$\pm$0.008
&  \bf{0.989$\pm$0.007}
&  0.928$\pm$0.021
&  0.988$\pm$0.004
&  0.998$\pm$0.001
\\
         DINOv2
&  0.868$\pm$0.023
&  0.973$\pm$0.010
&  0.988$\pm$0.006
& 0.865$\pm$0.023
&  0.975$\pm$0.005
&  0.997$\pm$0.002

\\
         BiomedCLIP
&  0.902$\pm$0.023
&  0.968$\pm$0.010
&  0.986$\pm$0.007
& 0.883$\pm$0.025
&  0.970$\pm$0.007
&  0.994$\pm$0.003

\\
         SAMMed-3D
&  0.739$\pm$0.047
&  0.924$\pm$0.017
&  0.967$\pm$0.008
& 0.609$\pm$0.068
&  0.838$\pm$0.018
& 0.937$\pm$0.009
\\
         MRI-CORE
&  0.810$\pm$0.034
&  0.943$\pm$0.017
&  0.973$\pm$0.013
& 0.720$\pm$0.044
&  0.900$\pm$0.014
&  0.975$\pm$0.005

\\
         BrainSegFounder
&  0.658$\pm$0.044
&  0.854$\pm$0.022
&  0.920$\pm$0.017
& 0.560$\pm$0.058
&  0.689$\pm$0.025
& 0.785$\pm$0.030

\\
\hline
        Decipher-MR Headneck
&  0.858$\pm$0.030
&  0.960$\pm$0.011
&  0.983$\pm$0.008
&  0.897$\pm$0.022
&  0.985$\pm$0.004
& 0.998$\pm$0.001
\\
         Decipher-MR T2
&  0.854$\pm$0.033
&  0.968$\pm$0.012
&  0.986$\pm$0.007
& 0.933$\pm$0.019
&  0.990$\pm$0.003
&  0.999$\pm$0.001
\\
         \hline
    \end{tabular}
    \begin{tabular}{l cccccccccccc}
    \hline
         &    \multicolumn{3}{c}{Contrast}&  \multicolumn{3}{c}{Motion}\\
         Models&  1\% (110) & 10\% (1100) & all (11008) &  10\% (34) & 25\% (87) & all (348)\\
         \hline
          Decipher-MR (high-reso)
& \bf{0.741$\pm$0.105}
&  \bf{0.898$\pm$0.049}
&  0.947$\pm$0.036
& 0.917$\pm$0.033
&  0.925$\pm$0.032
&  \bf{0.932$\pm$0.029}\\         \hline
          Decipher-MR
& 0.732$\pm$0.098
&  0.885$\pm$0.049
&  0.951$\pm$0.034
& 0.905$\pm$0.033
&  0.917$\pm$0.031
&  0.928$\pm$0.029\\
         Decipher-MR (1st stage)
& 0.723$\pm$0.102
&  0.883$\pm$0.047
&  \bf{0.952$\pm$0.035}
& 0.904$\pm$0.035
&  0.915$\pm$0.035
&  0.925$\pm$0.031
\\
        MedImageInsight
& 0.722$\pm$0.114
& 0.894$\pm$0.052
& 0.948$\pm$0.042
&  \bf{0.921$\pm$0.030}
& \bf{0.926$\pm$0.033}
& \bf{0.932$\pm$0.027}
\\
         DINOv2
&  0.695$\pm$0.083
& 0.856$\pm$0.057
& 0.936$\pm$0.044
&  0.910$\pm$0.033
& 0.916$\pm$0.032
& 0.924$\pm$0.032

\\
         BiomedCLIP
&  0.697$\pm$0.097
& 0.830$\pm$0.064
& 0.923$\pm$0.053
&  0.896$\pm$0.029
& 0.905$\pm$0.033
& 0.914$\pm$0.030

\\
         SAMMed-3D
&  0.592$\pm$0.095
& 0.715$\pm$0.063
& 0.830$\pm$0.058
&  0.563$\pm$0.064
& 0.607$\pm$0.078
&0.692$\pm$0.053

\\
         MRI-CORE
&  0.628$\pm$0.092
& 0.772$\pm$0.056
& 0.886$\pm$0.054
&  0.864$\pm$0.042
& 0.878$\pm$0.033
& 0.893$\pm$0.033

\\
         BrainSegFounder
&  0.526$\pm$0.081
& 0.530$\pm$0.106
& 0.752$\pm$0.072
&  0.541$\pm$0.070
& 0.569$\pm$0.066
&0.584$\pm$0.064

\\
\hline
        Decipher-MR Headneck
& 0.713$\pm$0.092
& 0.857$\pm$0.047
& 0.938$\pm$0.036
&  0.900$\pm$0.035
& 0.910$\pm$0.039
& 0.920$\pm$0.032
\\
         Decipher-MR T2
&  0.665$\pm$0.102
& 0.877$\pm$0.048
& 0.948$\pm$0.039
&  0.910$\pm$0.030
& 0.917$\pm$0.029
& 0.925$\pm$0.032
\\
         \hline
    \end{tabular}
    \caption{Performance on prediction of imaging-related attributes. \zy{Mean AUC and standard deviation across 50 random data splits are reported for all models and tasks. Results are shown at different low-data percentages, where only a subset of the training set is used to fine-tune the model, with the number of images for each subset indicated next to the percentage.}}
    \label{tab:classification-imaging}
\end{table}

\begin{table}[!ht]
    \centering
    \begin{tabular}{l ccccc c ccc}
    \hline
 & \multicolumn{5}{c}{ADNI}& \multicolumn{1}{c}{ACDC} & \multicolumn{3}{c}{LLD-MMRI}\\
         Models&  10\%&  25\%&  all& MCI& AD&   all&   10\%& 25\%& all\\
         \hline
         Decipher-MR (1\ts{st} stage)
&  \textcolor{black}{$1.9e$-3}
&  $4.2e$-1
&  $8.3e$-1
&  $3.5e$-2
&  $9.5e$-1
&  \textcolor{black}{$2.5e$-6}
&  \textcolor{black}{$1.8e$-4}
&  $2.8e$-1
&  $8.8e$-2
\\
         MedImageInsight
&  \textcolor{black}{$2.6e$-17}
&  \textcolor{black}{$8.8e$-12}
&  \textcolor{black}{$9.3e$-15}
&  \textcolor{black}{$3.2e$-4}
&  \textcolor{black}{$2.5e$-17}
&  \textcolor{black}{$3.0e$-3}
&  $3.5e$-1
&  \textcolor{black}{$1.2e$-2}
&  \textcolor{black}{$1.2e$-3}
\\
         \hline
    \end{tabular}
        \begin{tabular}{l ccc ccc}
    \hline
 & \multicolumn{3}{c}{PICAI}& \multicolumn{3}{c}{PICAI-Crop}\\
         Models& 10\%& 25\%& all&  10\%& 25\%& all\\
         \hline

         Decipher-MR (1\ts{st} stage)
&  $5.1e$-1
&  $6.2e$-1
&  $4.2e$-1
&  $2.2e$-1
&  $1.6e$-1
&  \textcolor{black}{$7.8e$-6}
\\
         MedImageInsight
&  $6.6e$-1
&  $6.5e$-1
&  $5.9e$-1
&  \textcolor{black}{$1.5e$-3}
&  \textcolor{black}{$8.7e$-6}
&  \textcolor{black}{$4.4e$-17}
\\
         \hline
    \end{tabular}
    \begin{tabular}{l cccccc ccc ccc}
    \hline
 & \multicolumn{6}{c}{ADNI}& \multicolumn{3}{c}{PICAI}& \multicolumn{3}{c}{PICAI-Crop} \\
         &  \multicolumn{3}{c}{Sex ($\uparrow$)}&  \multicolumn{3}{c}{Age ($\downarrow$)}&  \multicolumn{3}{c}{Age ($\downarrow$)}&  \multicolumn{3}{c}{Age ($\downarrow$)}\\
         Models&  10\% & 25\% & all &  10\% & 25\% & all &  10\% & 25\% & all&  10\% & 25\% & all\\
         \hline
         Decipher-MR (1\ts{st} stage)
&  \textcolor{black}{$1.6e$-7}
&  \textcolor{black}{$8.8e$-5}
&  \textcolor{black}{$2.6e$-2}
&  \textcolor{black}{$3.1e$-13}
&  \textcolor{black}{$2.3e$-5}
&  $8.9e$-2
& $5.5e$-1
&  $9.5e$-2
&  $8.5e$-2
&  \textcolor{black}{$1.3e$-4}
&  \textcolor{black}{$3.5e$-7}
&  \textcolor{black}{$4.7e$-9}
\\
         MedImageInsight
&  $8.2e$-1
&  $7.0e$-1
&  $3.1e$-2
&  \textcolor{black}{$4.4e$-21}
&  \textcolor{black}{$3.3e$-21}
&  \textcolor{black}{$2.8e$-19}
&  \textcolor{black}{$2.5e$-11}
&  \textcolor{black}{$2.8e$-12}
&  \textcolor{black}{$8.9e$-18}
&  \textcolor{black}{$2.8e$-6}
&  \textcolor{black}{$5.3e$-14}
&  \textcolor{black}{$1.4e$-21}
\\
         \hline
    \end{tabular}
    \begin{tabular}{l cccccccccccc}
    \hline
         &  \multicolumn{3}{c}{Body region}&  \multicolumn{3}{c}{Image sequence}&  \multicolumn{3}{c}{Contrast}&  \multicolumn{3}{c}{Motion}\\
         Models&  1\% & 10\% & all &  1\% & 10\% & all &  1\% & 10\% & all &  10\% & 25\% & all\\
 \hline
    
         Decipher-MR (1st stage)
&  \textcolor{black}{$9.2e$-25}
&  \textcolor{black}{$6.5e$-8}
&  $1.5e$-1
&  \textcolor{black}{$5.4e$-20}
&  \textcolor{black}{$3.8e$-11}
&  \textcolor{black}{$2.3e$-3}
& $1.6e$-1
& $5.2e$-1
&  $4.9e$-1
& $7.0e$-1
&  $3.3e$-1
&  $4.4e$-1
\\
        MedImageInsight
&  \textcolor{black}{$4.9e$-13}
&  $3.7e$-2
&  \textcolor{black}{$5.9e$-3}
&  \textcolor{black}{$5.3e$-3}
&  \textcolor{black}{$4.7e$-8}
&  \textcolor{black}{$1.3e$-4}
& \textcolor{black}{$1.3e$-4}
&  $4.6e$-1
&  $1.2e$-1
&  \textcolor{black}{$1.6e$-4}
& \textcolor{black}{$1.8e$-2}
& \textcolor{black}{$2.4e$-2}
\\
         \hline
    \end{tabular}
    \caption{\zy{Statistical comparisons on classification and regression tasks. P-values from paired t-tests comparing performances between Decipher-MR and MedImageInsight, as well as the vision-only variant, across 50 random splits. P-values significant under the Benjamini–Hochberg procedure at FDR 0.05 are highlighted in red.}}
    \label{tab:classification-pvalue}
\end{table}

\begin{table}[!ht]
    \centering
    \begin{tabular}{l cccccccccccc}
    \hline
         & \multicolumn{3}{c}{PICAI}&  \multicolumn{3}{c}{PICAI-Crop}\\
         LR & $1.0e$-3 & $1.0e$-4 & $1.0e$-5 & $1.0e$-3 & $1.0e$-4 & $1.0e$-5\\
 \hline
    
         Decipher-MR
& 0.698$\pm$0.043
&  0.700$\pm$0.040
&  0.699$\pm$0.047
& 0.762$\pm$0.041
&  0.770$\pm$0.039
&  0.763$\pm$0.041
\\
        MedImageInsight
& 0.713$\pm$0.042 
& 0.708$\pm$0.041
& 0.709$\pm$0.044
&  0.720$\pm$0.044
& 0.723$\pm$0.045
& 0.713$\pm$0.048
\\
         \hline
    \end{tabular}
    \begin{tabular}{l cccccccccccc}
    \hline
         &  \multicolumn{3}{c}{ADNI-MCI} &  \multicolumn{3}{c}{ADNI-AD}\\
         LR & $1.0e$-3 & $1.0e$-4 & $1.0e$-5 & $1.0e$-3 & $1.0e$-4 & $1.0e$-5 \\
 \hline
    
         Decipher-MR
&  0.609$\pm$0.042
&  0.615$\pm$0.037
&  0.612$\pm$0.041
&  0.822$\pm$0.046
&  0.822$\pm$0.046
&  0.828$\pm$0.041
\\
        MedImageInsight
&  0.584$\pm$0.049
&  0.589$\pm$0.043
&  0.590$\pm$0.050
&  0.751$\pm$0.049
&  0.757$\pm$0.048
&  0.766$\pm$0.049
\\
         \hline
    \end{tabular}
    \begin{tabular}{l cccccccccccc}
    \hline
         &  \multicolumn{3}{c}{LLDMMRI} &  \multicolumn{3}{c}{MRART-Motion}\\
         LR & $1.0e$-3 & $1.0e$-4 & $1.0e$-5 & $1.0e$-3 & $1.0e$-4 & $1.0e$-5 \\
 \hline
    
         Decipher-MR
&  0.765$\pm$0.038
&  0.775$\pm$0.036
&  0.776$\pm$0.040
&  0.926$\pm$0.029
&  0.928$\pm$0.029
&  0.924$\pm$0.030
\\
        MedImageInsight
&  0.783$\pm$0.040
&  0.788$\pm$0.037
&  0.777$\pm$0.040
&  0.931$\pm$0.031
&  0.932$\pm$0.027
&  0.929$\pm$0.031
\\
         \hline
    \end{tabular}
    \caption{\zy{Robustness of classification probing results across different learning rates. Mean AUC and standard deviation across 50 random data splits are reported}}
    \label{tab:classification-lr}
\end{table}

\begin{table}[H]
\centering
\begin{tabular}{lcccccc}
\hline
& \multicolumn{6}{c}{\zy{Hold-out test set of pretraining dataset}} \\
Query & \multicolumn{3}{c}{Entire Report} & \multicolumn{3}{c}{Conclusion} \\
Models & @5 & @10 & @20 & @5 & @10 & @20 \\
\hline
Decipher-MR & 0.176 & 0.258& 0.361&  0.105 & 0.170& 0.258 \\
BiomedCLIP  & 0.007&0.018&0.038& 0.011&  0.022&0.041 \\
MedImageInsight  & 0.030 & 0.051 & 0.085 & 0.025 & 0.040 & 0.074 \\
\hline
\end{tabular}
\caption{Zero-shot cross-modal retrieval \zy{on the Hold-out test set of pretraining dataset}}
\label{tab:dp188_retrieval}
\end{table}

\begin{table}[H]
\centering
\begin{tabular}{lccccccc}
\multicolumn{8}{c}{AMOS Per-Region Segmentation Results}\\
\hline
Region          & Spleen & Right kidney & Left kidney & Gallbladder & Esophagus & Liver\\ 
Dice & 0.961  & 0.960 & 0.962 & 0.846 & 0.770 &  0.973 \\\hline
Region          & Stomach & Aorta & Inferior vena cava & Pancreas & Right adrenal gland & Left adrenal gland & Duodenum\\ 
Dice         & 0.913  & 0.927 & 0.900 & 0.883 & 0.640 & 0.665 & 0.737 \\ \hline
\end{tabular}\\
\vspace{0.5em} 
\begin{tabular}{lccc}
\multicolumn{4}{c}{ACDC Per-Region Segmentation Results}\\
\hline
Region          & Left Ventricle & Right Ventricle & Myocardium \\ 
Dice & 0.890$\pm$0.010  & 0.883$\pm$0.008  & 0.937$\pm$0.006 \\
\hline
\end{tabular}\\
\vspace{0.5em} 
\begin{tabular}{lccccc}
\multicolumn{6}{c}{MRDLAS-Pelvis Per-Region Segmentation Results}\\
\hline
Region          & Bladder & Bowel bag & Femoral head left & Femoral head right & Pelvis body\\ 
Dice & 0.946$\pm$0.009  & 0.915$\pm$0.009 & 0.934$\pm$0.008 & 0.935$\pm$0.004 & 0.990$\pm$0.004 \\\hline
Region          & Penile bulb & Prostate & Rectum & Seminal Vesicles & Urethra \\ 
Dice         & 0.833$\pm$0.014 & 0.884$\pm$0.025 & 0.850$\pm$0.020 & 0.777$\pm$0.036 & 0.413$\pm$0.029 \\ \hline
\end{tabular}\\
\caption{\zy{Per-organ segmentation performance on three datasets. For ACDC and MRDLAS, we report the mean and standard deviation of Dice scores across five folds; for AMOS, we report only the mean Dice score on the official evaluation split.}}
\label{tab:segmentation-spacing}
\end{table}

\begin{table}[H]
\centering
\begin{tabular}{lcccc}
\hline
           & Median & 25\% & 10\% & Max \\ \hline
AMOS (Abdomen) & 0.801  & 0.857 & 0.851 & 0.851 \\
ACDC          & 0.869  & 0.879 & 0.879 & 0.908 \\ \hline
\end{tabular}
\caption{Segmentation performance across different resampling spacings. Mean Dice scores are reported.}
\label{tab:segmentation-spacing}
\end{table}

\begin{table}[H]
\centering
\begin{tabular}{lcccc}
\hline
           & Inference Time & GPU Memory utlization \\ \hline
Decipher-MR & 0.024 seconds  & 2574.77MB  \\
MedImageInsight          &  4.698 seconds & 21944.47MB \\ 
DINOv2          & 0.987 seconds  & 2565.56MB \\ 
BiomedCLIP          & 1.026 seconds  & 2081.15MB \\ 
SAMMed-3D          &  2.702 seconds & 19305.55MB \\ 
MRI-CORE          & 27.959 seconds  & 18581.28MB \\ 
\hline
\end{tabular}
\caption{\zy{Total inference time and peak GPU memory usage for embedding extraction on a 3D MRI volume (256×256×176) using a 24-GB NVIDIA A10G GPU. Methods unable to process the full volume in one pass run over multiple iterations, with total time reported.}}
\label{tab:segmentation-spacing}
\end{table}

\end{document}